\documentclass[conference]{IEEEtran}
\usepackage{times}

\usepackage[numbers]{natbib}
\usepackage{multicol}
\usepackage[bookmarks=true]{hyperref}

\usepackage{amsmath}
\usepackage{csquotes}   
\usepackage{graphicx}   
\usepackage{multirow}   
\usepackage{booktabs}   
\usepackage{wrapfig}    
\usepackage{subcaption} 
\usepackage{xcolor}
\usepackage{capt-of}
\usepackage{afterpage}
\usepackage{cleveref}
\usepackage{amsfonts}
\crefname{figure}{Figure}{Figures}

\newcommand{\norm}[1]{\left\lVert#1\right\rVert}

\definecolor{vysics_purple}{HTML}{7030A0}
\definecolor{bsdf_blue}{HTML}{4472C4}
\definecolor{pll_green}{HTML}{70AD47}

\newcommand{\edit}[1]{{\textcolor{purple}{#1}}}

\begin{document}

\title{Vysics: Object Reconstruction Under Occlusion by Fusing Vision and Contact-Rich Physics}


\author{\authorblockN{Bibit Bianchini\authorrefmark{1}\authorrefmark{2}, Minghan Zhu\authorrefmark{1}\authorrefmark{2}, Mengti Sun\authorrefmark{3}, Bowen Jiang\authorrefmark{2}, Camillo J. Taylor\authorrefmark{2}, Michael Posa\authorrefmark{2}}
\authorblockA{\authorrefmark{1}The first two authors contributed equally to this work.}
\authorblockA{\authorrefmark{2}General Robotics, Automation, Sensing, and Perception (GRASP) Laboratory\\ University of Pennsylvania, Philadelphia, PA 19104\\ Emails: {\tt\small \{bibit, minghz, bwjiang, cjtaylor, posa\}@seas.upenn.edu}}
\authorblockA{\authorrefmark{3}Amazon, Seattle, WA 98004 \\ Email: {\tt\small mengtis@amazon.com}}}



%

\maketitle

\begin{abstract}
    We introduce Vysics, a vision-and-physics framework for a robot to build an expressive geometry and dynamics model of a single rigid body, using a seconds-long RGBD video and the robot’s proprioception. While the computer vision community has built powerful visual 3D perception algorithms, cluttered environments with heavy occlusions can limit the visibility of objects of interest. However, observed motion of partially occluded objects can imply physical interactions took place, such as contact with a robot or the environment. These inferred contacts can supplement the visible geometry with “physible geometry,” which best explains the observed object motion through physics. Vysics uses a vision-based tracking and reconstruction method, BundleSDF, to estimate the trajectory and the visible geometry from an RGBD video, and an odometry-based model learning method, Physics Learning Library (PLL), to infer the “physible” geometry from the trajectory through implicit contact dynamics optimization. The visible and “physible” geometries jointly factor into optimizing a signed distance function (SDF) to represent the object shape. Vysics does not require pretraining, nor tactile or force sensors. Compared with vision-only methods, Vysics yields object models with higher geometric accuracy and better dynamics prediction in experiments where the object interacts with the robot and the environment under heavy occlusion.  Project page:  \href{https://vysics-vision-and-physics.github.io/}{https://vysics-vision-and-physics.github.io/}
\end{abstract}

\IEEEpeerreviewmaketitle


\section{Introduction} \label{sec:intro}
\begin{figure}[t]
    \centering
    \begin{subfigure}[b]{0.48\columnwidth}
        \centering
        \includegraphics[width=\columnwidth]{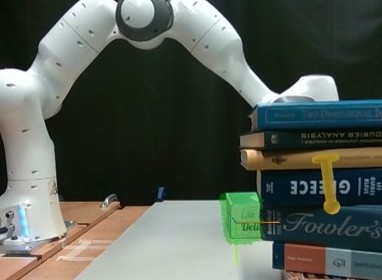}
        \caption{BundleSDF \cite{wen2023bundlesdf} }
        \label{fig:overview_bsdf}
    \end{subfigure}
    \hfill
    \begin{subfigure}[b]{0.48\columnwidth}
        \centering
        \includegraphics[width=\columnwidth]{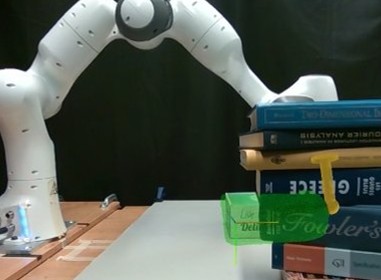}
        \caption{Vysics (ours) }
        \label{fig:overview_vysics}
    \end{subfigure}
    \caption{Vision-based shape reconstruction (projection shown in green) can be limited by occlusion. Fusing vision and contact-rich physics, our method recovers the occluded geometry through object interactions with the robot and environment. The robot end effector in yellow shows the robot-object interaction. }
    \label{fig:overview}
\end{figure}

In-the-wild manipulation will require robots to encounter a vast array of different objects.  While some might be recognized from an existing database, others will require physical interaction to be newly understood on the spot.  Dexterous manipulation of these objects will benefit from the ability to rapidly learn or identify object properties: geometry is most critical, but inertial properties are also valuable for predicting motion, particularly under forceful manipulation.
Use of such models boasts the benefits of interpretability and expected generalizability, at the cost of requiring the model.  This paper presents Vysics, which builds dynamics models of novel objects from vision and physical interaction, even in the face of substantial visual occlusions (Figure \ref{fig:overview}).

Rapid modeling requires combining all available information in a unified fashion.  This work leverages recent results from visual tracking and object reconstruction \cite{wen2023bundlesdf} in combination with contact-implicit model learning \cite{bianchini2023simultaneous, pfrommer2020contactnets} via the shared connection of object geometry
Depth camera measurements on an object's surface are direct observations of portions of its geometry,
and observations of the object's state evolution can inject more geometric information when contact is inferred.  Visual information is limited by occlusions, but contact, which typically occurs on visually-occluded faces of objects, provides a secondary source of information:  \enquote{physible} geometry.

Consider the example depicted in Figure \ref{fig:overview} of a robot arm interacting with an object, which is significantly obscured by a stack of books in the foreground.  A vision-only approach would only reconstruct the portions of the geometry that is visible but is also able to detect object motions.  By reasoning about physics, we can infer more about the geometry by assuming object trajectories are explainable by its physical interactions with the environment and robot.  In the Figure \ref{fig:overview} example, the robot end effector is occluded as it pushes the object from its occluded end.  While not explained by visible geometry, this guides estimation of the \enquote{physible} geometry, namely that the geometry should extend to the right to make contact with the robot.

Estimating geometry through contact-rich interactions is not a trivial problem.  Object trajectories result from the sum of all forces on the object, including continuous and contact forces, whose interplay relies on more object properties than just geometry \cite{bianchini2023simultaneous}.  Additionally, making and breaking contact introduces multi-modality into the dynamics, further complicating system identification \cite{antonova2023rethinking, parmar2021fundamental}.  We take an approach that embraces the multi-modal nature of the dynamics \cite{bianchini2022generalization, pfrommer2020contactnets}, starting by feeding it visually-estimated trajectories, then fusing visually-observed with physically-inferred geometry.

Our aim is to leverage vision and contact-rich physics to perform dynamics model building directly from RGBD data in extreme low data regimes -- learning a dynamics model whose geometry matches or outperforms vision-based approaches while learning other critical simulation parameters like inertia, with only seconds of data.



\subsection{Contributions and Outline}
We make the following contributions in this work:
\begin{itemize}
    \item Introduce Vysics, a novel method that builds expressive dynamics models of objects from RGBD videos featuring contact-rich trajectories.  With seconds of data, Vysics can identify an object's geometry with no fundamental priors as well as its inertial properties, automatically generating an accompanying Unified Robotics Description Format (URDF) file and mesh.
    \item Present and make available a new RGBD video dataset of a Franka Panda robot arm interacting with objects on a table in the face of significant visual occlusions (code and data can be found on our \href{https://vysics-vision-and-physics.github.io}{project page}).
    \item Demonstrate our method's efficacy on shape reconstruction and learning dynamics models.  We compare against vision-only shape reconstruction and show the benefits of the additional contact physics information source with small amounts of data.
\end{itemize}

In \S \ref{sec:related}, we review related work from computer vision and dynamics model learning. We discuss preliminaries for our work in \S \ref{sec:background} before outlining Vysics in \S \ref{sec:approach}.  \S \ref{sec:experimental_setup} describes experiment details before \S \ref{sec:results} presents results, followed by a discussion of limitations in \S \ref{sec:limitations} and conclusion in \S \ref{sec:conclusion}.


\section{Related Work} \label{sec:related}

\subsection{Vision-Based Geometry Reconstruction and Completion}
Contact is at the heart of robotic manipulation, and contact is driven by geometry.  Computer vision has made strides in geometry reconstruction from image inputs. 
Dense visual SLAM systems \cite{bloesch2018codeslam, teed2021droid, mccormac2018fusion++} recover dense geometry from multi‑view RGB or depth streams, yet leave occluded regions unresolved. Neural implicit methods represent shapes as continuous fields: foundational works like \cite{park2019deepsdf, mescheder2019occupancy} encode object‑level geometry,  
while \cite{azinovic2022neural, murez2020atlas} extend these representations to full scenes. 

When portions of the object remain unseen, learned completion methods infer missing geometry from partial data. Synthetic shape datasets \cite{chang2015shapenet} enable learning shape priors of common object categories, facilitating various tasks including shape completion \cite{yan2022shapeformer, chu2023diffcomplete}, point cloud completion \cite{yu2021pointr, chen2023anchorformer}, and single-image shape reconstruction \cite{mittal2022autosdf, cheng2023sdfusion}. However, most operate in an object‑centered canonical frame, limiting their applicability to raw sensor data, with a few exceptions \cite{thai20213dsdfnet, shin2018pixels} leveraging 2.5D representations to guide camera-frame shape learning. Robotics and VR applications have prompted camera-frame shape completion from partial point clouds \cite{van2020learningpointsdf, mohammadi20233dsgrasp} or RGBD observations \cite{wu2023multiviewmcc, lionar2023numcc}, with recent extensions to multi‑object scenes \cite{wright2024vprism, iwase2024zerooctmae}. Moreover, advances in image generation models \cite{rombach2022highstablediffusion}, 3D scene representations \cite{mildenhall2021nerf, kerbl20233dgs}, and large-scale 3D object datasets \cite{deitke2023objaverse, deitke2023objaversexl} have spurred 3D generative pipelines \cite{liu2023zero123, melas2023realfusion, hong2023lrm, liu2024one2345++, xu2024grm, zou2024triplane}, 
though these typically require unobstructed RGB inputs and do not integrate depth.
Unlike these vision‑only and data‑driven alternatives, Vysics requires no pretraining or canonical pose assumptions: it relies solely on vision-based measurements and physics-based hypotheses extracted from a short RGBD video to regress a class-agnostic shape.

\subsection{Vision-Based Object Pose Estimation}
While there are many works that perform robotic manipulation without explicitly estimating object poses (e.g. \cite{chen2023visual}),  
access to object pose estimates remains a commonplace assumption in robotics.  Classical vision-based pose estimators require the 3D shape model of the target object to facilitate the pose estimate \cite{labbe2022megapose, labbe2020cosypose, park2019pix2pose}, often impractical in realistic applications.  Other new approaches do not require geometry models beforehand \cite{wen2023foundationpose}, but this has its own challenges.  Model-free methods without this assumption rely on sparse keypoints, dense optical flow, or deep features with reference frames with known pose \cite{castro2023posematcher, sun2022onepose}, without building a coherent object model.  Such approaches are susceptible to long-term drift, though even weak geometry-based supervision can help \cite{hou2020mobilepose}.  Limiting a pose estimator's scope to tracking in-category objects can boost pose accuracy \cite{lin2022keypoint, di2022gpv, chen2020category}, but limits generalization.

\subsection{Simultaneous Tracking and Shape Reconstruction}
Recent computer vision works combine the pose and shape estimation problems \cite{zhi2024simultaneous, jiang2024forge, wen2023bundlesdf, sucar2020nodeslam}.  Solving both problems simultaneously has its benefits, including that maintaining a geometry estimate can improve novel-object pose estimation and vice versa \cite{stoiber2022iterative, wang20206}.  Other applications include multiple sparse-view alignment \cite{liu2022gen6d, sun2022onepose, labbe2020cosypose} and single-view pose estimation \cite{nguyen2024nope, zhang2022relpose}.

\subsection{Trajectory-Based Dynamics Model Learning}
System identification is an important robotics subfield that aims to build accurate system models, which can then be leveraged via model-based control techniques.  Differentiable simulators \cite{le2023single, howell2022dojo, de2018end} have pushed recent advancements, though they can struggle to find good solutions when applied in contact-rich settings \cite{antonova2023rethinking, bianchini2022generalization}.  While high-stiffness contact dynamics generally are a challenge for system identification \cite{parmar2021fundamental}, more creative strategies can make finding inertial parameters \cite{fazeli2017learning}, contact parameters \cite{pfrommer2020contactnets}, or both \cite{bianchini2023simultaneous} possible and efficient.  Many of these methods rely on access to state information, limiting their use to lab-based settings where fiducials are available, or necessitating integration with a generalizable vision-based solution.

\subsection{Physics as a Prior for Shape Reconstruction}
Some prior works have used physics as a prior for shape reconstruction.  Unlike Vysics, these other works either assume objects of interest are statically stable \cite{ni2024phyrecon, agnew2021amodal} or avoid learning other non-geometric dynamic properties like inertia \cite{abou-chakra2024physically, song2018inferring}.  The most similar in spirit to Vysics is \cite{song2018inferring}, as they reconstruct geometries with occlusions through physical robot and environment interactions.  While \cite{song2018inferring} avoids the problematic gradients in contact-rich scenarios by using a gradient-free search over a discrete set of hypothesized geometries, Vysics leverages smooth, implicit-based losses and thus can directly regress the object geometry based on visual measurements.  This enables Vysics to avoid the need for a high-dimensional graph search that limits geometries to a relatively small set of possibilities.

\section{Background} \label{sec:background}

Vysics leverages a vision-based pose and shape estimation tool, BundleSDF \cite{wen2023bundlesdf}, and a physics-inspired dynamics learning tool, Physics Learning Library (PLL) \cite{bianchini2023simultaneous, pfrommer2020contactnets}, which are described in \S \ref{subsec:bundlesdf_geom} and \S \ref{subsec:pll_geom}, respectively.  
Neither of these tools requires pretraining on an extensive dataset.  BundleSDF and PLL both train on and generate results from only the measurements provided or inferred by their per-instance input data, making them suitable for immediate application when a robot encounters and needs to model an object.

\subsection{Vision-Based Pose and Shape Estimation}  
\label{subsec:bundlesdf_geom}
BundleSDF \cite{wen2023bundlesdf} takes masked RGBD videos as input, and outputs both estimated poses of the object over time as well as a geometry estimate.  It is built on top of the generalizable, model-free 6D pose tracker, BundleTrack \cite{wen2021bundletrack}.  Alongside the pose estimator, BundleSDF introduces a parallel Neural Object Field that simultaneously maintains an estimate of the object's geometry, which is used to further refine the estimated poses to encourage long-term consistency.

To represent geometry, BundleSDF uses a signed distance function (SDF), an expressive implicit shape representation with no imposed restrictions such as an inherent resolution limit or convexity prior.  Due to their expressivity, SDFs have seen great success in representing shapes of objects and environments, even with generalization to unseen objects \cite{park2019deepsdf}.  An SDF is a function of the form,
\begin{align}
    \text{SDF}(\mathbf{p}) &= d, \label{eqn:sdf}
\end{align}
where any 3D point $\mathbf{p}$ relative to a geometry's reference frame can be queried, and the scalar signed distance $d$ away to the nearest surface point is returned.  See Figure \ref{fig:geometry} for illustration.

From RGBD data, not only do depth returns imply points at which $d = 0$, but samples along the camera rays of valid object depth returns can additionally yield supervision for points with $d > 0$.  The result is that BundleSDF trains its neural network SDF with supervision from sets of points and signed distances $\{ (\mathbf{p}, d)_i \}$.  BundleSDF implements a hybrid SDF model on this set of points, by only regressing points whose target signed distance is within a truncation distance, as in \cite{suresh2023neural, park2019deepsdf}.  We refer interested readers to \cite{wen2023bundlesdf} for more details on BundleSDF loss components.  Upon termination, the marching cubes algorithm extracts a mesh from the underlying SDF.

\subsection{Physics-Inspired Dynamics Learning}
\label{subsec:pll_geom}

\begin{figure}
    \centering
    \includegraphics[width=0.9\linewidth]{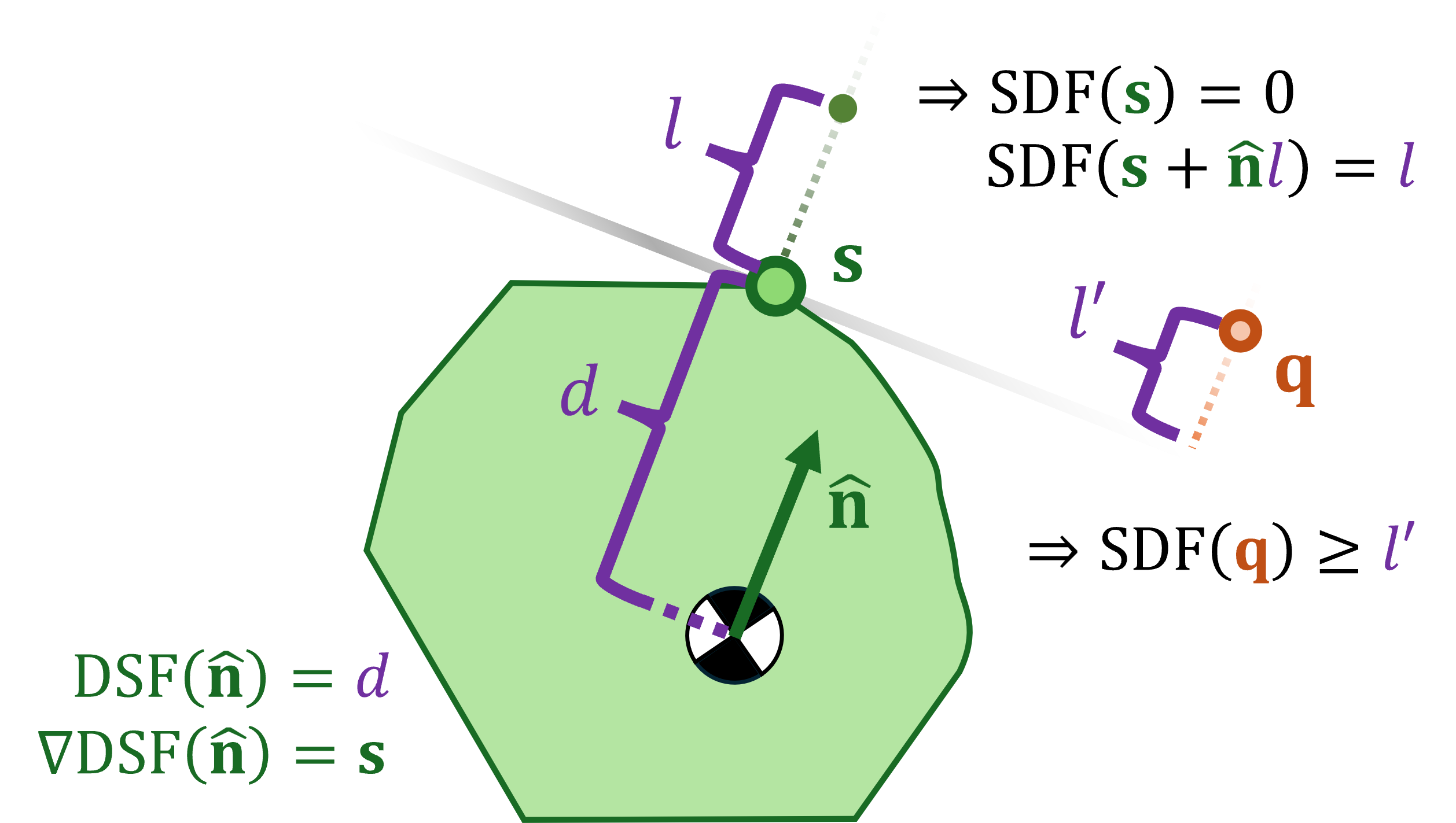}
    \caption{A 2D depiction of the physical meaning of a DSF \eqref{eqn:dsf} and its implication on the SDF \eqref{eqn:sdf}.  Shades of green points have exact SDF values and are subject to \eqref{eqn:loss_support_point}, and the orange example point $\mathbf{q}$'s signed distance can be lower-bounded by the supporting hyperplane as in \eqref{eqn:loss_hyperplane}.}
    \label{fig:geometry}
\end{figure}

PLL \cite{bianchini2023simultaneous, pfrommer2020contactnets} is an odometry-based method that jointly learns continuous and contact dynamics of objects undergoing contact-rich trajectories.  Given only a state trajectory (no direct observation of contact events), PLL is able to identify the geometry, frictional properties, and inertia, all from hypothesizing contact forces that best explain observed state transitions.  Its key insight is in its implicit loss formulation, which has provably better generalization than alternatives that require differentiable simulation \cite{bianchini2022generalization}.  The loss applies a cost to the current model belief's incompatibility with the hypothesized contact forces; see \cite{bianchini2023simultaneous, pfrommer2020contactnets} for more details.

The geometry of the object directly determines when it makes and breaks contact with the robot and environment, thus playing a critical role in its dynamics.  PLL solves the inverse problem:  estimate the contact geometry from observed dynamics. The estimated geometry is represented using PLL's deep support function (DSF) \cite{halm2023addressing}, an input-convex, homogeneous deep neural network \cite{amos2017input}.  Like an SDF, a DSF has no inherent resolution limits, however it can only represent the convex hull of any given shape.  In our application where our robot interacts with convex and nonconvex objects, this assumption is satisfied as long as collisions occur only on the object's convex hull.
A DSF has all the expressivity required for those scenarios while also being concise and fast to compute.

\begin{figure*}[t]
    \centering
    \includegraphics[width=\textwidth]{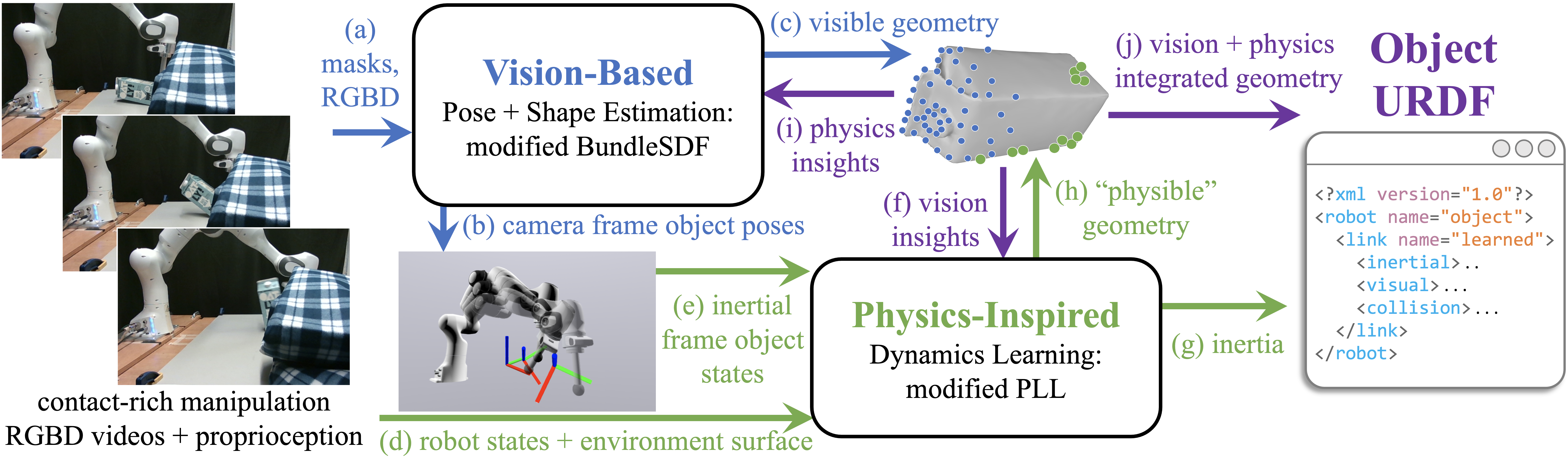}
    \caption{Detailed Vysics diagram.  \textcolor{bsdf_blue}{Blue} arrows denote the vision-based information flow through BundleSDF \cite{wen2023bundlesdf}, and \textcolor{pll_green}{green} arrows for PLL \cite{bianchini2023simultaneous, pfrommer2020contactnets}.  \textcolor{vysics_purple}{Purple} arrows indicate the unifying connections Vysics makes to factor both vision and contact-rich physics into the geometry learning problem.}
    \label{fig:cycle}
\end{figure*}

A DSF's input can be any 3D point, but for physically meaningful outputs, PLL only queries the DSF with unit vectors.  A DSF takes as input a unit vector in the object's body frame and outputs the scalar distance the geometry extends in that direction \cite{bazaraa2013nonlinear}.  The gradient of a DSF with respect to its input is (almost always \cite{halm2023addressing}) the 3D point on the object geometry that extends furthest in the queried input direction.  Mathematically, for an object whose surface (or volume) is represented by the set $\mathcal{S}$, a DSF yields the following output and gradient,
\begin{align}
    \text{DSF}(\mathbf{\hat{n}}) &= \max_{\mathbf{s}_i \in \mathcal{S}} \, \mathbf{s}_i \cdot \mathbf{\hat{n}}, \\ \nabla_\mathbf{\hat{n}} \text{DSF} \left( \mathbf{\hat{n}} \right) &= \arg \max_{\mathbf{s}_i \in \mathcal{S}} \, \mathbf{s}_i \cdot \mathbf{\hat{n}} =: \mathbf{s}. \label{eqn:dsf}
\end{align}
The queried normal direction $\mathbf{\hat{n}}$ and its associated support point $\mathbf{s}$ fully define a supporting hyperplane of the object geometry \cite{boyd2004convex}.  See Figure \ref{fig:geometry} for an example of a queried normal direction $\mathbf{\hat{n}}$, its corresponding support point $\mathbf{s}$, and the constraints they impose on the SDF (explained in detail in \cref{subsec:contact_in_vision_geom}).  A mesh may be exported by PLL with vertices $\{ \mathbf{s}_i \}$ as outputs of $\nabla_\mathbf{\hat{n}} \text{DSF}$ on a uniformly sampled set of unit vectors $\{ \mathbf{\hat{n}}_i \}$.


\section{Approach} \label{sec:approach}

Figure \ref{fig:cycle} illustrates Vysics from input RGBD videos and robot states (left) to URDF output (right).  Its core components are BundleSDF \cite{wen2023bundlesdf} for vision-based tracking and shape reconstruction, and PLL \cite{bianchini2023simultaneous, pfrommer2020contactnets} for physics-inspired dynamics learning. 
Beyond the insights that led to this systems integration, our main contribution lies in how Vysics incorporates these two powerful tools together such that they supervise each other and output an object dynamics model, featuring geometry informed by both vision and contact.

Referring to the labeled arrows in Figure \ref{fig:cycle}, we obtain the object trajectory (b) and the initial shape estimates (c) from masked input RGBD images (a) via BundleSDF.  The object trajectories are converted to an inertial reference frame, where the table surface (d) identified in the depth images is on a known plane.  From this plane (d) and the processed poses (e), PLL detects \enquote{physible} portions of the geometry by inferring contact events in the observed dynamics.  PLL is also subject to supervision from BundleSDF (f) to encourage consistency with the visible geometry.  BundleSDF runs again, fusing both the visible (a) and \enquote{physible} data (i) into a geometric model consistent with both information streams.  The final output of Vysics inherits the physics-supervised inertial parameters (g) and jointly-supervised geometry (j).  This is exported as a URDF which can be simulated to produce dynamics predictions.  

The basis of our contribution is in how we unify the visible and \enquote{physible} geometry measurements together.  \S \ref{subsec:vision_in_contact_geom} discusses how vision helps in the contact learning problem (f), and \S \ref{subsec:contact_in_vision_geom} describes how we inject the geometric information estimated from the contact dynamics back into the vision-based shape estimation (i).

\subsection{Supervising Contact-Based Geometry with Vision} \label{subsec:vision_in_contact_geom}
Contact-based geometry reconstruction has the ability to learn the full convex hull of any shape as long as every point on its convex hull makes contact with the known environment plane or robot over its recorded trajectory.  In extreme low-data regimes where many sides of the geometry never contact the robot or ground, the complete geometry is unobservable, and the solutions of the geometry that can accurately explain the trajectories are non-unique.  Since PLL's violation-based implicit loss function has terms that scale with the magnitude of inferred contact forces \cite{pfrommer2020contactnets}, it is encouraged to find the DSF that minimizes the inferred contact force magnitudes while explaining the trajectory data.  In finding a geometry-force combination that explains a net observed torque, PLL can be biased to learn large geometries to increase lever arms and thus reduce force magnitudes.  Without intervention, this scenario is common and problematic in low-data, where many sides of the object may never contact the robot or ground. However, vision can provide an additional source of information (denoted by superscript $^v$) to disambiguate the possible geometries that may explain the trajectory. 

Given an RGBD video, BundleSDF yields an object mesh corresponding to the zero SDF level set. We first calculate the convex hull of the mesh since DSF in PLL can only represent convex shapes. Then, we apply a visibility check to preserve only the vertices visible in the RGBD video, and we call this set of points $\mathcal{V}$, the visible geometry. 

The visible geometry $\mathcal{V}$ can supervise the surface predicted by PLL's DSF.  For each $\mathbf{s}^v\in \mathcal{V}$, we \edit{wish to} penalize the distance from the nearest surface point $\mathbf{s}^p$ predicted by the DSF,
\begin{align}
    \frac{1}{|\mathcal{V}|} \sum_{\mathbf{s}^{v}\in \mathcal{V}} \norm{\mathbf{s}^p - \mathbf{s}^v}. \label{eqn:vision_loss}
\end{align}
However, it is not straightforward to locate the $\mathbf{s}^p$ surface point on the DSF closest to a given point, since the DSF is an implicit shape representation.  To approximate, we sample the DSF to create a mesh and find the point $\mathbf{s}^{p'}$ on its surface that is closest to $\mathbf{s}^{v}$. The vector
\begin{equation}
    \mathbf{\hat{n}}^{p'}=\frac{\mathbf{s}^v-\mathbf{s}^{p'}}{\norm{\mathbf{s}^v-\mathbf{s}^{p'}}}
\end{equation}
is an approximation of the querying vector $\mathbf{\hat{n}}^{p}$ such that $\nabla \text{DSF}(\mathbf{\hat{n}}^{p}) = \mathbf{s}^{p}$. 
The approximation is up to the angular resolution of the unit vector samples when converting the DSF to a mesh.  The direction of the vector $\mathbf{\hat{n}}^{p'}$ is chosen to always point outwards from the shape.  In the end, we use 
\begin{equation}\label{eqn:vision_loss_actual}
    \mathcal{L}_\text{bsdf} = \frac{1}{|\mathcal{V}|} \sum_{\mathbf{s}^{v}\in \mathcal{V}} \norm{\nabla\text{DSF}(\mathbf{\hat{n}}^{p'}) - \mathbf{s}^v}
\end{equation}
as the vision-based supervision of the PLL training. See \cref{fig:bsdf2pll} for an illustration. See Appendix \ref{apx:hyperparams} for hyperparameters, including the relative loss term weights in PLL. 


\begin{figure}[t]
    \centering
    \begin{subfigure}[b]{0.41\columnwidth}
        \centering
        \includegraphics[width=\columnwidth]{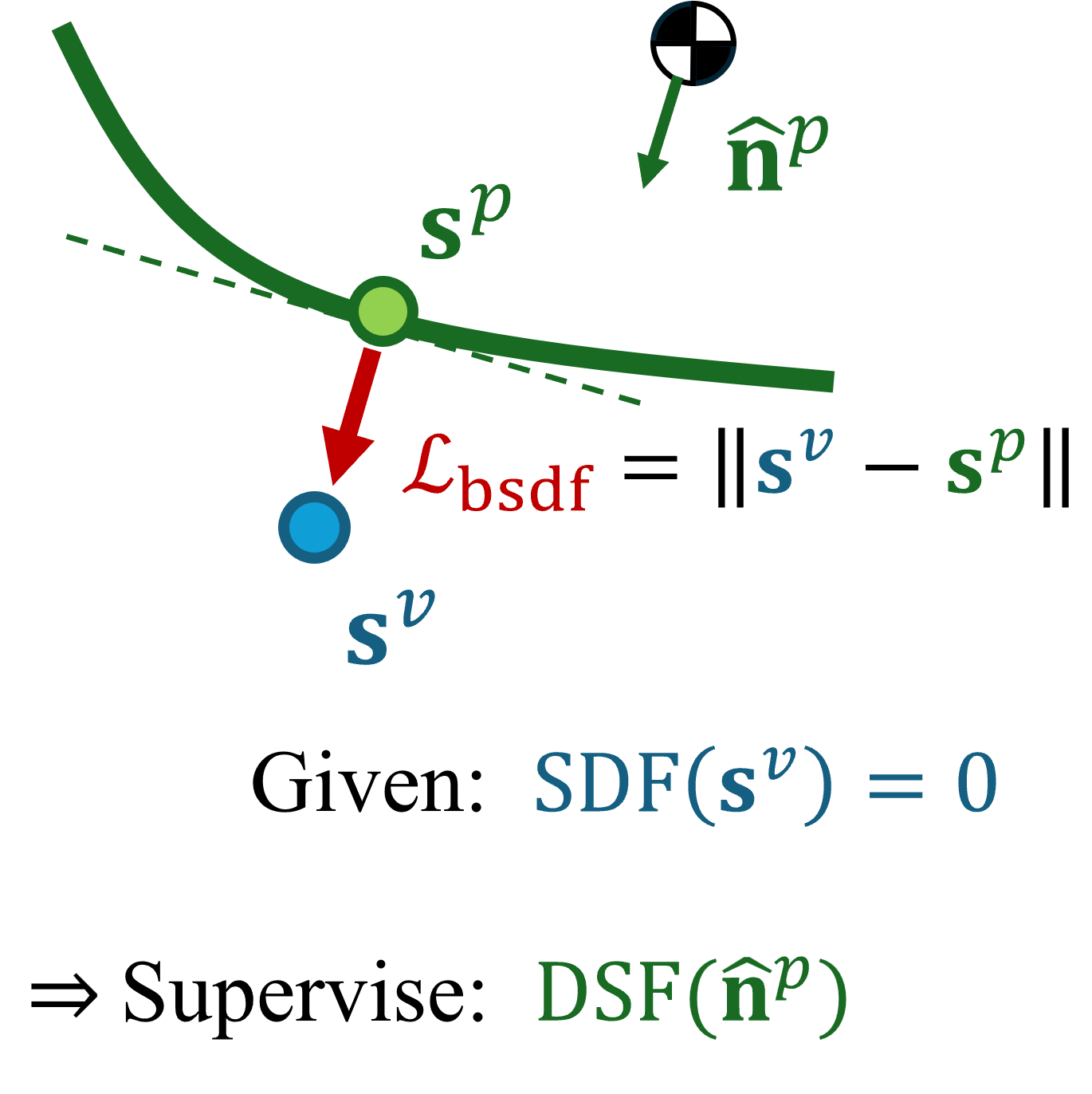}
        \caption{Supervising contact-based DSF network with visible geometry $\mathbf{s}^v\in \mathcal{V}$. }
        \label{fig:bsdf2pll}
    \end{subfigure}
    \hfill
    \begin{subfigure}[b]{0.55\columnwidth}
        \centering
        \includegraphics[width=\columnwidth]{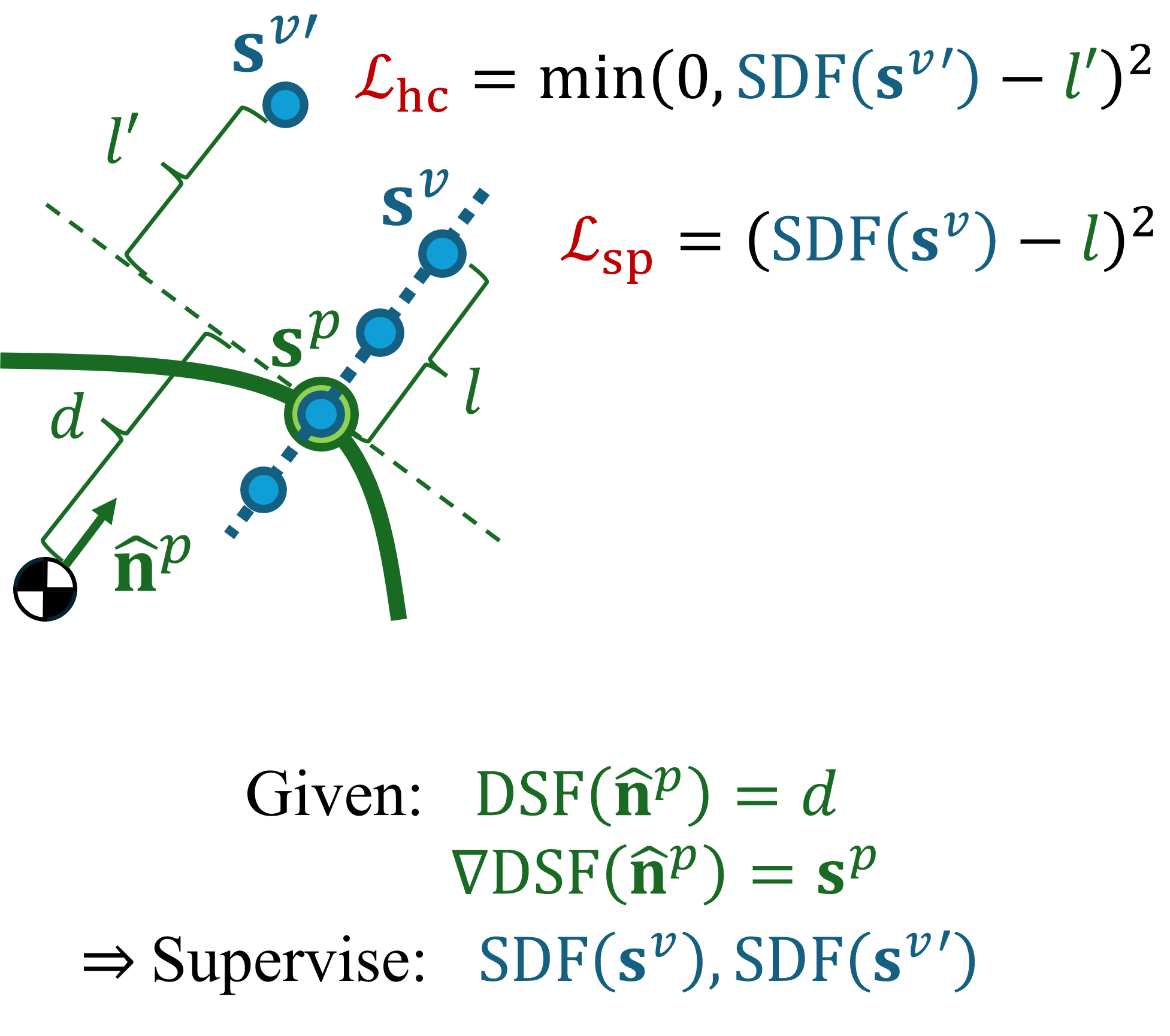}
        \caption{Supervising vision-based SDF network with \enquote{physible} geometry $(\mathbf{\hat{n}}^p, \mathbf{s}^p)\in \mathcal{P}$. }
        \label{fig:pll2bsdf}
    \end{subfigure}
    \caption{Visualization of the loss functions as the incorporation of vision and contact dynamics. Blue represents the geometry learned from vision, and green represents the geometry learned from contact dynamics. }
    \label{fig:loss_interplay}
\end{figure}

\subsection{Supervising Vision-Based Geometry with Contact}\label{subsec:contact_in_vision_geom}

As mentioned in \cref{subsec:vision_in_contact_geom}, the SDF estimated by BundleSDF may only capture the visible geometry, while PLL estimates a convex shape, in the form of DSF, that best explains the trajectory through contact dynamics. We combine the two streams of information into a single geometry estimation. We use SDF as the final geometry representation, given its capability to model non-convex shapes. The SDF training in the BundleSDF is run on the RGBD video for the second time but with new contact-induced losses from the DSF, as introduced below.

\subsubsection{From PLL to the \enquote{Physible} Geometry}

The DSF from PLL can be represented as a set of $\nabla \text{DSF}$ input/output pairs, 
$\{(\mathbf{\hat{n}}^p, \mathbf{s}^p)\}$.
However, only the local area where contacts happened may be effectively supervised through contact dynamics optimization, while the geometry of other areas is ambiguous with unreliable DSF value. Therefore, we filter the set and only preserve an $(\mathbf{\hat{n}}^p, \mathbf{s}^p)$ pair if it is associated with contact force, estimated by PLL, above a certain threshold. The filtered set, $\mathcal{P}$,
is the \enquote{physible} geometry output from PLL. 


\subsubsection{Support Point Loss}
Given a \enquote{physible} geometry data point $(\mathbf{\hat{n}}^p, \mathbf{s}^p)$, not only is $\mathbf{s}^p$ on the object boundary and should have $\text{SDF}(\mathbf{s}^p)=0$, but we can use $\mathbf{\hat{n}}^p$ to infer constraints on the local geometric region.  Consider the ray $\vec{r}$ from $\mathbf{s}^p$ pointing in direction $\mathbf{\hat{n}^p}$.  Any point on this ray, lying a distance $l \in [0, \infty)$ on the ray from $\mathbf{s}$, has a signed distance of exactly $l$,
\begin{align}
    \text{SDF} \left( \mathbf{s}^p + l \mathbf{\hat{n}}^p \right) &= l.
\end{align}
See \cref{fig:geometry} for illustration.  For $\mathbf{s}^v = \mathbf{s}^p + l \mathbf{\hat{n}}^p$, we denote $\overline{\text{SDF}}(\mathbf{s}^v):= l$ to represent the signed distance induced from DSF.  If we extend the possible range for $l$ values to go negative, i.e. $l \in [-\epsilon, \infty)$, then we can encourage zero-crossings in the SDF network, but the negative signed distance values may not be strictly correct for certain geometries and choices of $\epsilon$.  In practice, this can be accurate enough during supervision with small $\epsilon$ and enables more meaningful change at the geometry surface.  Thus, we introduce a \textbf{support point loss} for $\mathbf{s}^v \in \mathcal{P}_{\text{sp}}(\mathbf{\hat{n}}^p, \mathbf{s}^p)$,
\begin{equation}\label{eqn:loss_support_point}
    \mathcal{L}_\text{sp} = \left( \overline{\text{SDF}}(\mathbf{s}^v) - \text{SDF}(\mathbf{s}^v) \right)^2, 
\end{equation}
where
\begin{equation}\label{eqn:support_points}
    \mathcal{P}_{\text{sp}}(\mathbf{\hat{n}}^p, \mathbf{s}^p)=\{\mathbf{s}^v | \mathbf{s}^v = \mathbf{s}^p + l \mathbf{\hat{n}}^p, \ \ l \in [-\epsilon, \infty)\}
\end{equation} 
is the set of \textit{support points} induced by the \enquote{physible} point $(\mathbf{\hat{n}}^p, \mathbf{s}^p)$. $\mathcal{L}_\text{sp}$ encourages signed distance consistency at and around observed \enquote{physible} points.  Figure \ref{fig:pll2bsdf} depicts four such sampled points $\mathbf{s}^v$ and their supervised signed distance values $l$ induced from a single support point $\mathbf{s}^p$.

\subsubsection{Hyperplane-Constrained Loss}
On this ray, \eqref{eqn:loss_support_point} imposes strong supervision on the SDF network, though only local to areas near PLL-inferred contacts.  However, as depicted in \cref{fig:geometry}, \textit{any} point $\mathbf{q}$ can have its signed distance minimum-bounded based on a support direction/point pair $(\mathbf{\hat{n}}, \mathbf{s})$.  Consider a pair $(\mathbf{\hat{n}}^p, \mathbf{s}^p)\in \mathcal{P}$, its supporting hyperplane   implies that the signed distance at $\mathbf{s}^{v'}$ can be lower bounded by the distance from $\mathbf{s}^v$ to the supporting hyperplane,
\begin{align}
    \text{SDF}(\mathbf{s}^{v'}) &\geq \left( \mathbf{s}^{v'} - \mathbf{s}^p \right) \cdot \mathbf{\hat{n}}^p.
\end{align}
Thus, we introduce a \textbf{hyperplane-constrained loss} valid for any $\mathbf{s}^{v'} \in \mathbb{R}^3$,
\begin{align}
    \mathcal{L}_\text{hc} &= \min \left( 0, \, \text{SDF}(\mathbf{s}^{v'}) - ( \mathbf{s}^{v'} - \mathbf{s}^p ) \cdot \mathbf{\hat{n}}^p \right) ^2.  \label{eqn:loss_hyperplane}
\end{align}
Figure \ref{fig:pll2bsdf} gives one example of a point $\mathbf{s}^{v'}$ whose signed distance is lower bounded with respect to a support direction/point \eqref{eqn:loss_support_point}. While $\mathbf{s}^{v'}$ may be sampled arbitrarily, we sample them around a cylindrical neighborhood of the support points in practice. 

\subsubsection{Convexity Loss}
In comparison to the dense visible points, \enquote{physible} points are sparser and can be located far away from the set of visible points.  With the assumption that the robot is interacting with a single object at a time, we add a bias loss term to encourage the estimated shape to be convex when no observed RGBD data signals otherwise.  This helps the sparse contact points attach to the visible shape in the SDF regression. We randomly sample pairs of points from the near-surface points estimated by the SDF network, $\mathbf{S}^{v_0}=\{\mathbf{s}^v\in \mathbb{R}^3 \ | \ | \text{SDF}(\mathbf{s}^v)| \leq \epsilon \}$, and the collection of support points \eqref{eqn:support_points} induced by all \enquote{physible} points $\mathbf{S}^{p_0}=\{\mathbf{s}^v | \mathbf{s}^v\in \mathbf{S}_{\text{sp}}(\mathbf{\hat{n}}^p, \mathbf{s}^p), (\mathbf{\hat{n}}^p, \mathbf{s}^p) \in \mathcal{P} \}$. 
Points in $\mathbf{S}^{v_0}$ have SDF supervision from the RGBD video, while points in $\mathbf{S}^{p_0}$ have induced $\overline{\text{SDF}}$ supervision from the contact dynamics. 
Then we sample an interpolated point between each pair of points. The interpolated SDF serves as the upper bound of the SDF prediction at the interpolation point if we assume the shape is convex, 
\begin{equation}
    \text{SDF}(\mathbf{s}^v) \leq l \ \text{SDF}(\mathbf{s}_1) + (1-l) \ \overline{\text{SDF}}(\mathbf{s}_2),
\end{equation}
for $\mathbf{\mathbf{s}}^v = l \mathbf{s}_1 + (1-l) \mathbf{s}_2, 0\leq l \leq 1$, where $\mathbf{s}_1 \in \mathbf{S}^{v_0}, \mathbf{s}_2\in \mathbf{S}^{p_0}$ are samples from the near-surface points and support points. Correspondingly, the \textbf{convexity loss} for point $\mathbf{q}$ is,
\begin{equation}
    \mathcal{L}_\text{cvx} = \max \left( 0, \text{SDF}(\mathbf{\mathbf{s}}^v ) - \left( l \ \text{SDF}(\mathbf{s}_1) + (1-l) \ \overline{\text{SDF}}(\mathbf{s}_2) \right) \right).  \label{eqn:loss_convex}
\end{equation}
See Appendix \ref{apx:hyperparams} for hyperparameters used in our experiments, including the relative loss term weights in BundleSDF.




\section{Experimental Setup} \label{sec:experimental_setup}

\subsection{Dataset}
We consider a new dataset of RGBD videos of a Franka Emika Panda arm with a spherical end effector interacting with an object repeatedly on a flat table surface.  These robot interactions were teleoperated via commanded end effector poses tracked with impedance control. The dataset includes the RGBD videos of the objects in interactions with object mask annotations, as well as the robot joint states.  In PLL, the object can collide with the tip of the end effector (modeled as a sphere) and the table surface (modeled as a plane).  The end-effector sphere location is known from the robot joint states in combination with the transform between the camera and the robot base, and the height of the table surface is detected automatically based on the depth readings from the camera.  The ground truth shapes of the objects are provided in the form of meshes for evaluation.  The end-effector pose commands by the teleoperator are also included for the dynamics prediction evaluation (see \cref{sec:metrics}).  There are substantial visual occlusions preventing the camera from directly seeing much of the object geometry.   

\begin{figure} 
   \centering
    \includegraphics[width=0.9\linewidth]{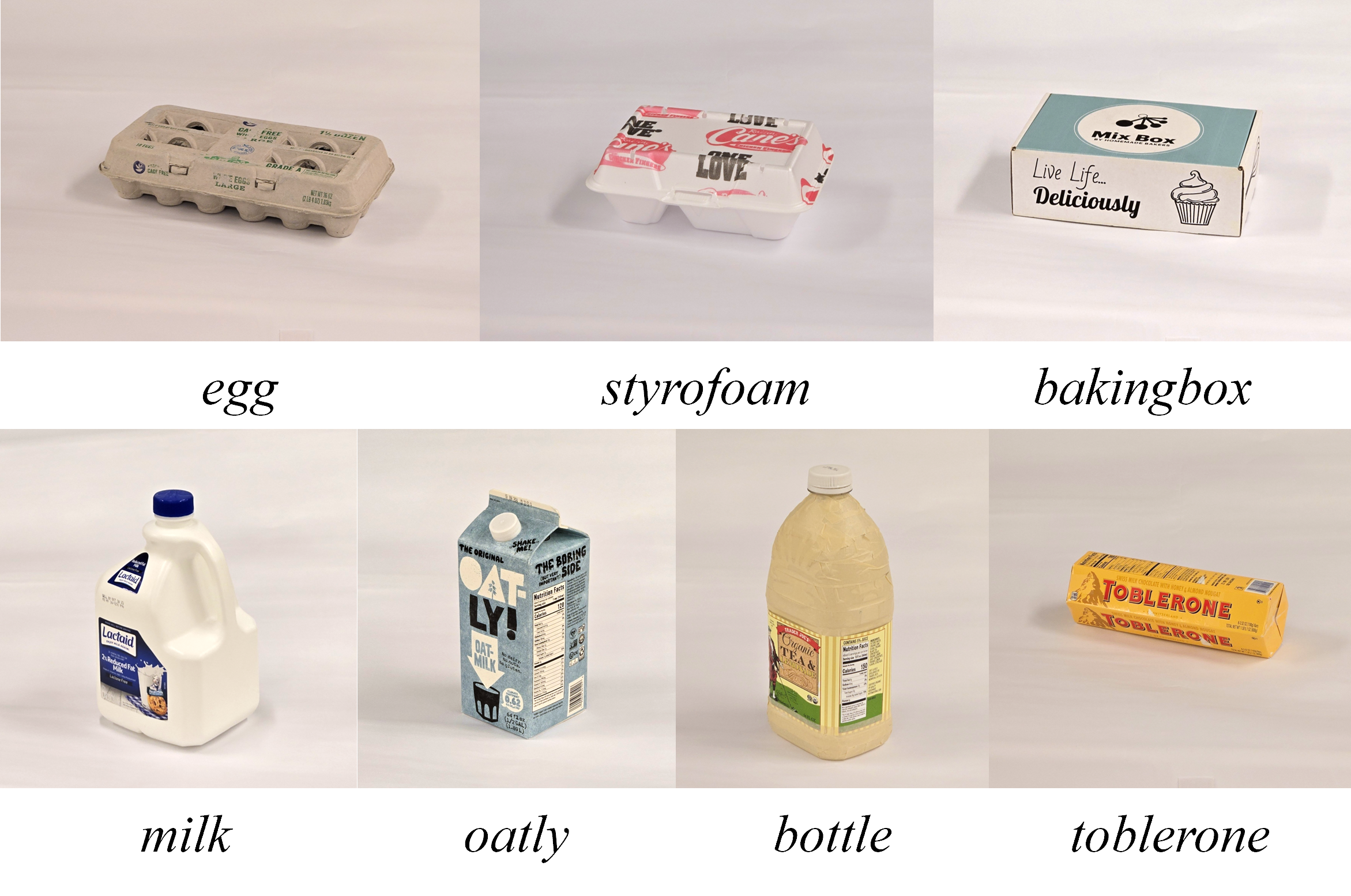}
    \caption{The 7 objects and their names in our dataset.  }
    \label{fig:objects}
\end{figure}

The objects in our dataset are a mixture of everyday items depicted in Figure \ref{fig:objects}.  
We used a RealSense D455 collecting 640x480 pixel RGBD images at 30Hz.  The object masks for a video were semi-automatically generated from manual masks on the first frame using XMem \cite{cheng2022xmem}. For every object, we collected multiple sessions of the robot arm interacting with the object with its the spherical end effector. Each session is around 10 seconds long.  The dataset varies in starting configurations, foreground occlusions, and robot interactions, e.g. sliding vs. pivoting vs. toppling and in what directions.  In the evaluation, we excluded the sessions in which BundleSDF lost track of the object and failed to yield the object trajectory. Therefore, the number of sessions for each object vary.


While PLL is capable of identifying friction coefficients essentially by observing acceleration during periods of sliding, sliding motions in our dataset largely occur during sustained object-robot contact, making it difficult to uniquely identify frictional ground contact forces from contact forces with the robot.
Thus, we use PLL to learn only geometry and inertia.  We set the pair-wise friction coefficients to reasonable values of 0.26 (object-table) and 0.15 (object-robot) for all experiments, higher for object-table because of a high-friction silicone mat on the table.

\subsection{Metrics}\label{sec:metrics}
We evaluate the performance of the identified geometry and dynamics in two ways. First, we evaluate the geometric error between the estimated shape and the ground truth shape. Specifically, we report the chamfer distance and the volumetric intersection-over-union (IoU) between the learned geometry and the ground truth geometry,
which we align with manual annotation and ICP for refinement.  This alignment step is only for geometric evaluation, not for training or deployment of the learned model.  Second, we perform dynamics predictions to evaluate the geometry through the resulting trajectories.  We implemented our impedance controller in simulation and ran it on the recorded end effector pose commands to reproduce the robot's actions. Using the estimated object geometry and its tracked pose at the first frame as the initial condition, we generate simulated trajectories of the object. The predicted trajectories are compared with the real-world trajectory tracked by BundleSDF to show how well the estimated geometry explains the dynamics.

To quantify the dynamics prediction error, we use three metrics.  The first is \textit{average pose error} (split into position and rotation terms), which measures the difference between the tracked pose and the simulated pose throughout the open-loop simulation. 
As contact dynamics are naturally chaotic, any open-loop simulation is subject to substantial uncertainty.
To provide an alternative assessment of the accuracy of the learned model, 
our second metric is \textit{time-before-divergence}, which measures the time from the start of the simulation to the point when the simulated pose error is higher than a given threshold. We use a positional threshold of 10 cm and a rotational threshold of 45 degrees.  The third is the \textit{temporal IoU of contact activation}, which measures the overlap of when the robot-object contact happens in the simulated rollout and in the real experiment. The real contact duration is manually annotated by inspecting the RGBD videos. This metric highlights the compatibility between the physical contact and the estimated geometry.  Both average pose error and time before divergence are with respect to the BundleSDF estimated poses.

\begin{table}[]
    \centering
    \setlength{\tabcolsep}{3pt}
    \resizebox{\columnwidth}{!}{
    \begin{tabular}{lccccccc|c}
    \toprule
    Method    & bakingbox & bottle & egg  & milk & oatly & styrofoam & toblerone & all \\ \midrule
    3DSGrasp \cite{mohammadi20233dsgrasp}  & 3.83      & 2.80   & 3.78 & 3.15 & 2.51  & 2.66      & 2.77      & 3.06    \\
    IPoD \cite{wu2024ipod}     & 3.25      & 1.80   & 2.16 & 2.37 & 2.73  & 1.93      & 1.97      & 2.47    \\
    V-PRISM \cite{wright2024vprism} & 3.52 & 2.47 & 2.31 & 3.33 & 2.30 & 2.54 & 2.48 & 2.80 \\
    OctMAE \cite{iwase2024zerooctmae}   & 3.11      & 2.22   & 1.52 & 2.93 & 2.13  & 2.00      & 2.36      & 2.45    \\ \midrule
    BundleSDF \cite{wen2023bundlesdf} & 3.84      & 2.65   & 3.70  & 3.17 & 2.45  & 2.55      & 2.44      & 2.98    \\
    \textit{Vysics (ours)}      & \textbf{1.83}      & \textbf{1.36}   & \textbf{1.05} & \textbf{1.53} & \textbf{1.25}  & \textbf{1.45}      & \textbf{1.02}      & \textbf{1.45}   \\ \bottomrule
    \end{tabular}
    }
    \caption{Average chamfer distance (unit: cm) of shape completion baselines compared with BundleSDF and our method.}
    \label{tab:chamfer_baselines}
\end{table}

\begin{figure}[t]
    \centering
    \begin{subfigure}[b]{0.48\columnwidth}
        \centering
        \includegraphics[width=0.65\columnwidth]{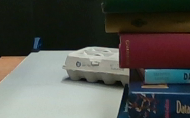}
        \caption{RGB input}
        \label{fig:gen_rgb}
    \end{subfigure}
    \hfill
    \begin{subfigure}[b]{0.48\columnwidth}
        \centering
        \includegraphics[width=0.4\columnwidth]{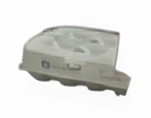}
        \caption{OpenLRM \cite{hong2023lrm,openlrm} }
        \label{fig:gen_lrm}
    \end{subfigure}
    \hfill
    \begin{subfigure}[b]{0.48\columnwidth}
        \centering
        \includegraphics[width=0.4\columnwidth]{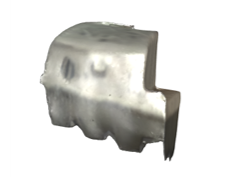}
        \caption{One-2-3-45++ \cite{liu2024one2345++}}
        \label{fig:gen_12345}
    \end{subfigure}
    \hfill
    \begin{subfigure}[b]{0.48\columnwidth}
        \centering
        \includegraphics[width=0.4\columnwidth]{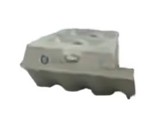}
        \caption{TriplaneGaussian \cite{zou2024triplane}}
        \label{fig:gen_trig}
    \end{subfigure}
    \caption{A qualitative example of generative single-view reconstruction on an occluded RGB image of the \textit{egg} object.}
    \label{fig:gen_qual}
\end{figure}

\begin{figure}[!t]
    \centering
    \subfloat[An example reconstruction of \textit{bottle} in a pushing interaction.  ]{\includegraphics[width=\columnwidth]{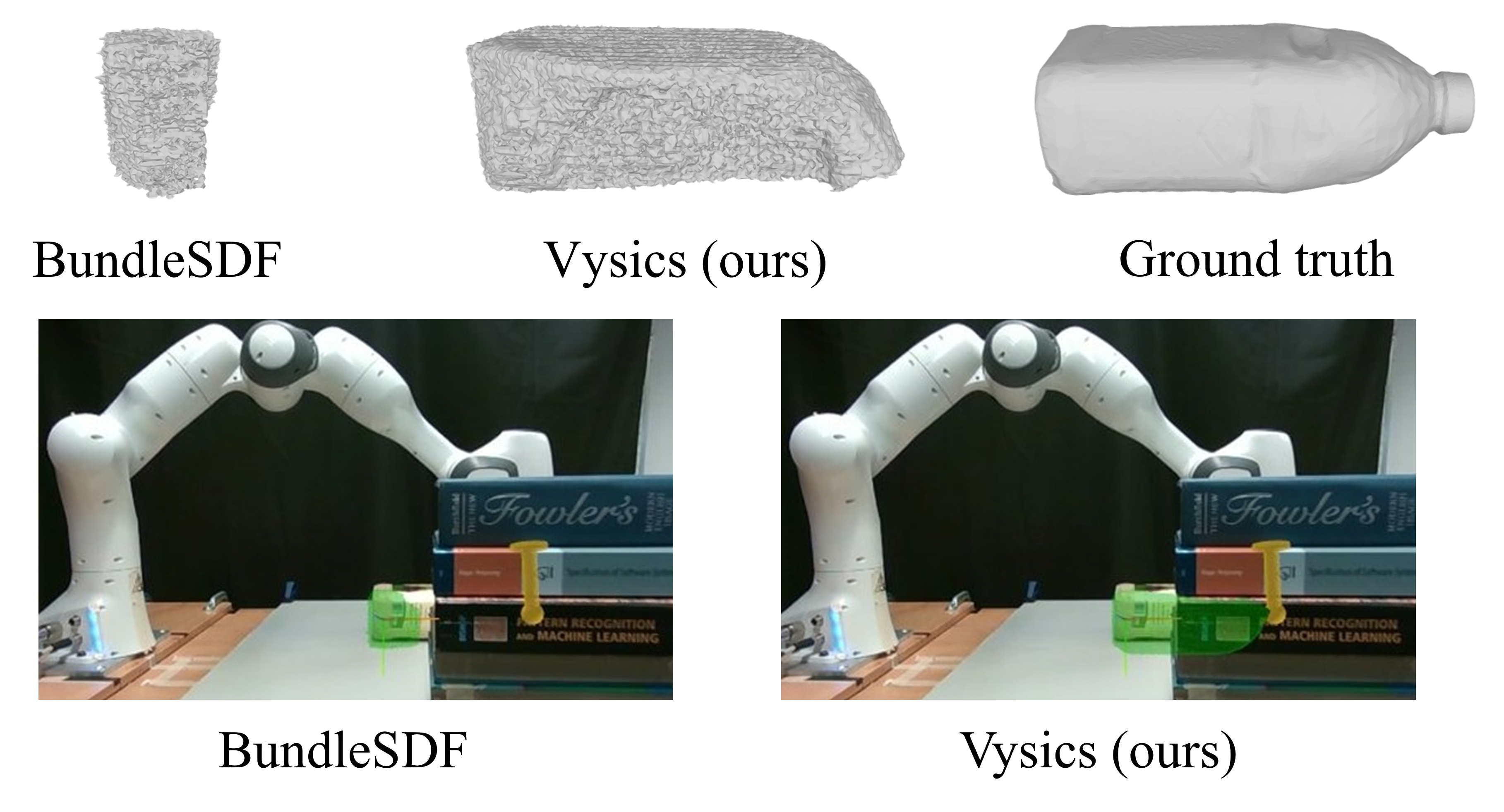}
    \label{fig:subfig_bottle_1}}
    \vspace{10pt}
    \subfloat[An example reconstruction of \textit{oatly} in a pivoting interaction. ]{\includegraphics[width=\columnwidth]{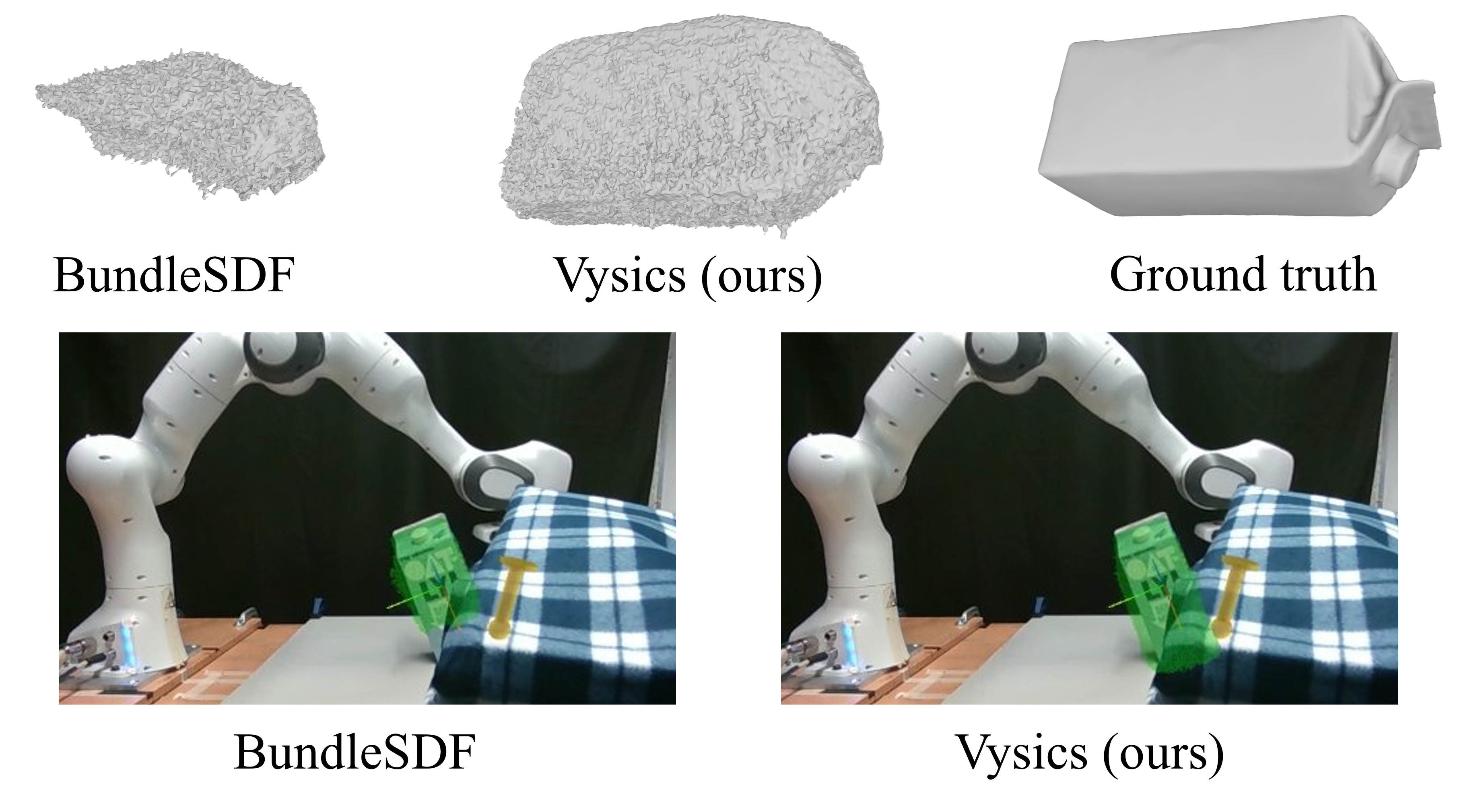}
    \label{fig:subfig_oatly_6}}
    
    \caption{A qualitative comparison of the geometry reconstruction under heavy occlusion between our method and the vision-only baseline. In the image view, the mesh projection is shown in green, and the robot end effector is shown in yellow to illustrate the robot-object interaction. }
    \label{fig:geometry_qualitative}
\end{figure}

\begin{figure}[t]
    \centering
    \includegraphics[width=\linewidth]{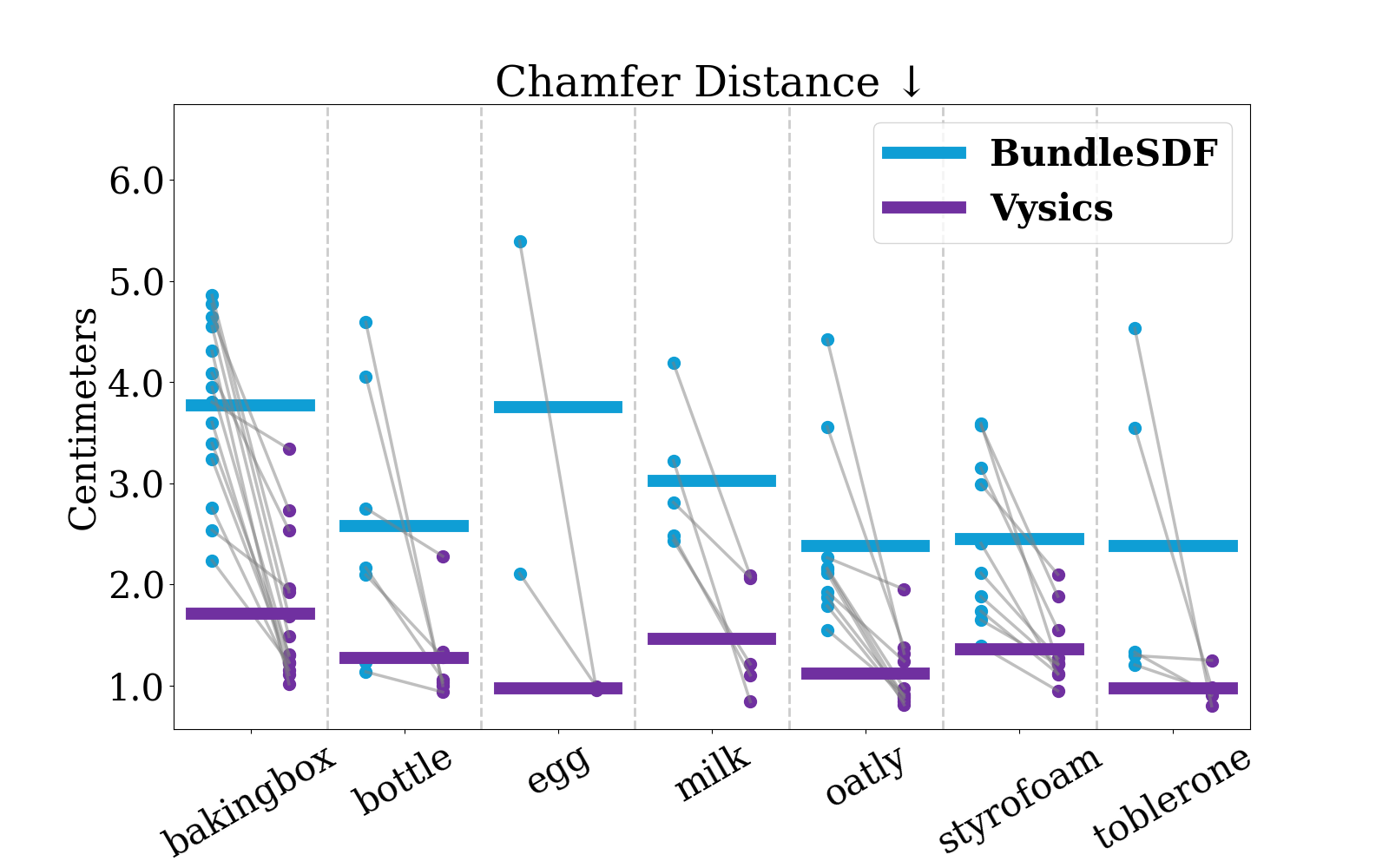}
    \includegraphics[width=\linewidth]{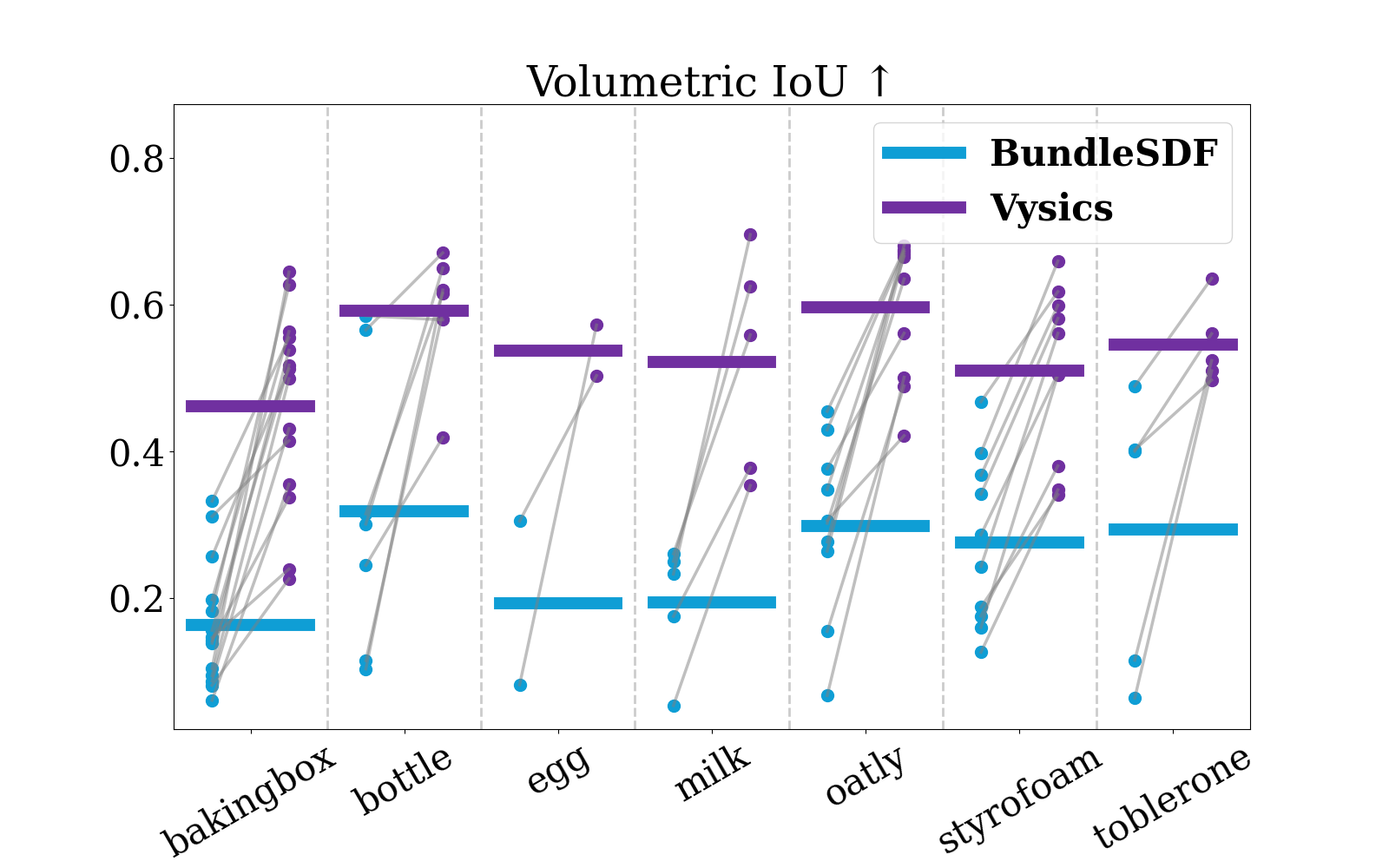}
    \caption{The quantitative comparison of the geometric reconstruction accuracy. Each dot is one session. The results of the same session from different methods are connected by a gray line. $\uparrow$ means higher is better. $\downarrow$ means lower is better. }
    \label{fig:geo_quantitative}
\end{figure}

\subsection{Baselines}
Using quantitative geometry reconstruction metrics, we compare against vision-only methods, including BundleSDF and a variety of shape completion methods:  3DSGrasp \cite{mohammadi20233dsgrasp} for point cloud completion (we use points from visible mesh faces reconstructed by BundleSDF as input); IPoD \cite{wu2024ipod} for single-object completion from an RGBD image; and V-PRISM \cite{wright2024vprism} and OctMAE \cite{iwase2024zerooctmae} for multi-object scene completion from an RGBD image.  All of these methods, like Vysics, are category-agnostic.  For shape completion methods based on an RGBD image, we use the last frame of each video clip, which, in our data, is typically less occluded. We provide object masks for IPoD and foreground masks for V-PRISM and OctMAE, then segment the object from their scene completion outputs. We use published pretrained models for evaluation.
We additionally provide a qualitative comparison 
against methods of 3D reconstruction from a single RGB image:  OpenLRM \cite{openlrm, hong2023lrm}, One-2-3-45++ \cite{liu2024one2345++}, and Triplane Gaussian \cite{zou2024triplane}, to see if these models trained with large-scale 2D and 3D data can recover the complete shape from a partial view.

For dynamics predictions, we compare against fair variations of BundleSDF and PLL.  The BundleSDF variation features the geometry from BundleSDF and inertia centered on its geometric centroid.  We explored instead using the inertia learned by PLL in combination with the BundleSDF geometry, but this often results in unrealistic dynamics predictions due to the PLL-learned center of mass landing outside the BundleSDF geometry's convex hull.  The PLL variation is the result of running PLL on the BundleSDF pose estimates plus vision supervision via \eqref{eqn:vision_loss_actual}; this is the same as Vysics without the second round of BundleSDF.



\begin{figure}[t]
    \centering
    \includegraphics[width=\linewidth]{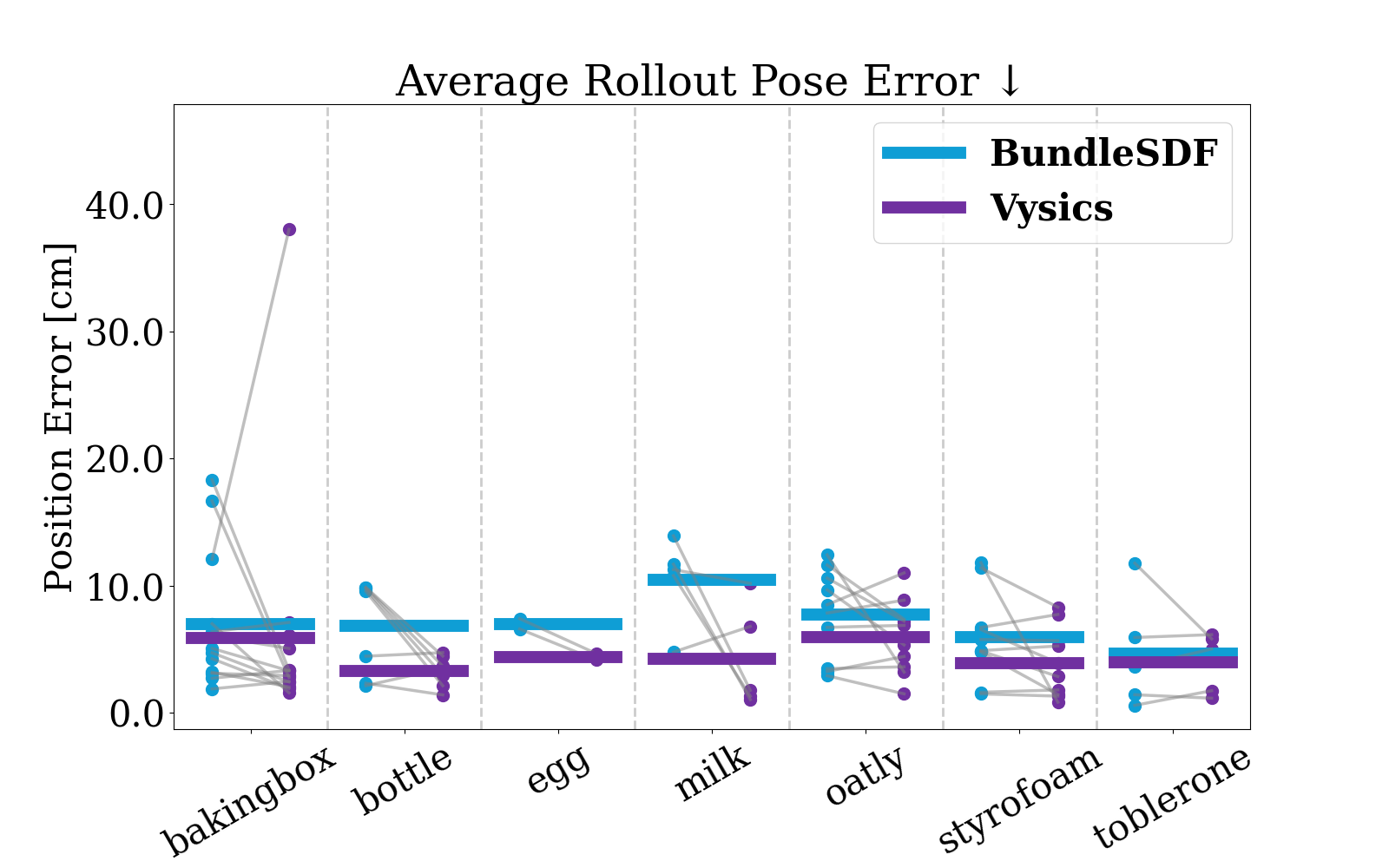}
    \includegraphics[width=\linewidth]{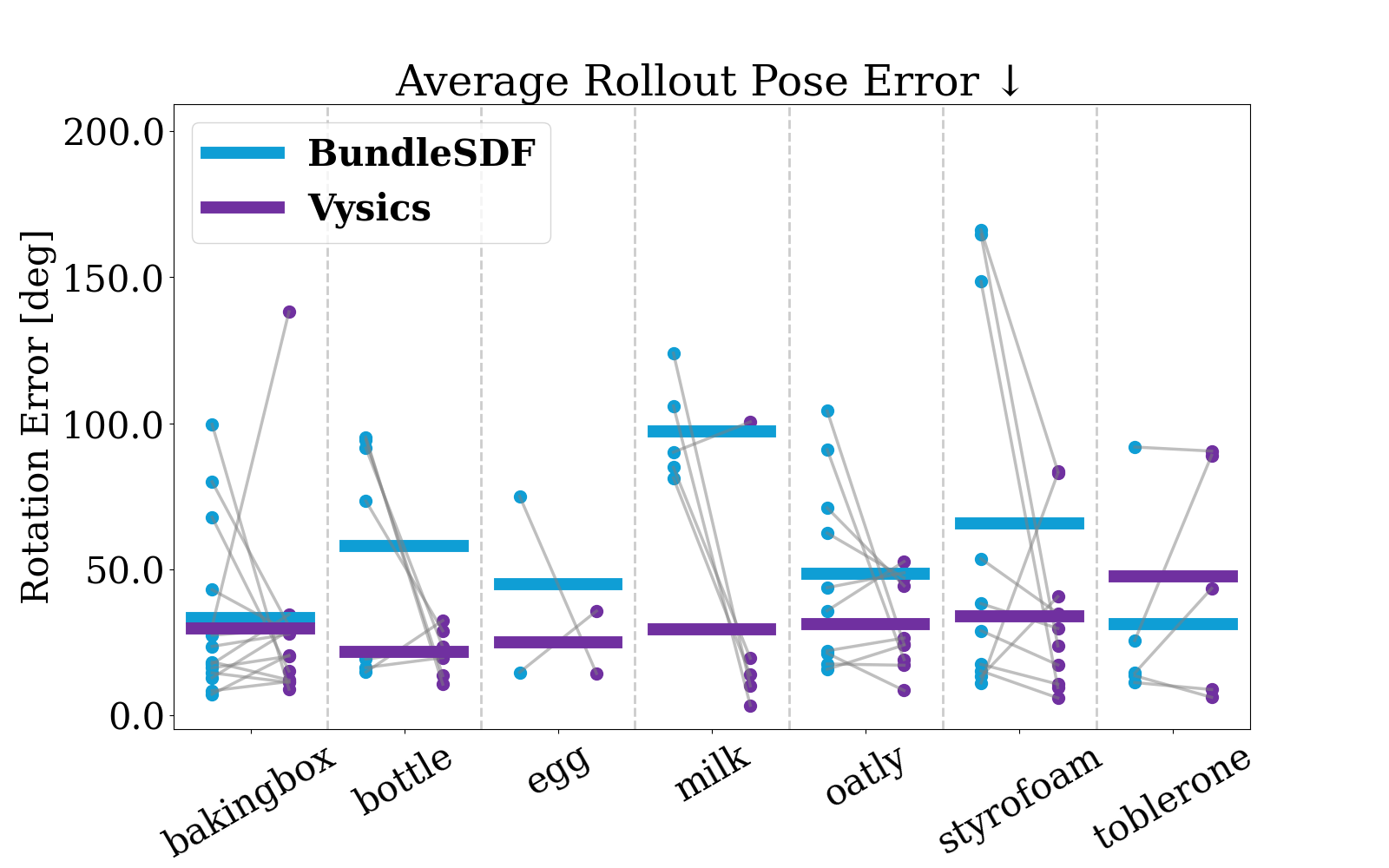}
    \caption{The quantitative comparison of the dynamics prediction accuracy in pose error. Trajectories are predicted by replaying the robot interaction with the estimated geometry in simulation. }
    \label{fig:dynamics_pose_error}
\end{figure}

\begin{figure}[t]
    \centering
    \includegraphics[width=\linewidth]{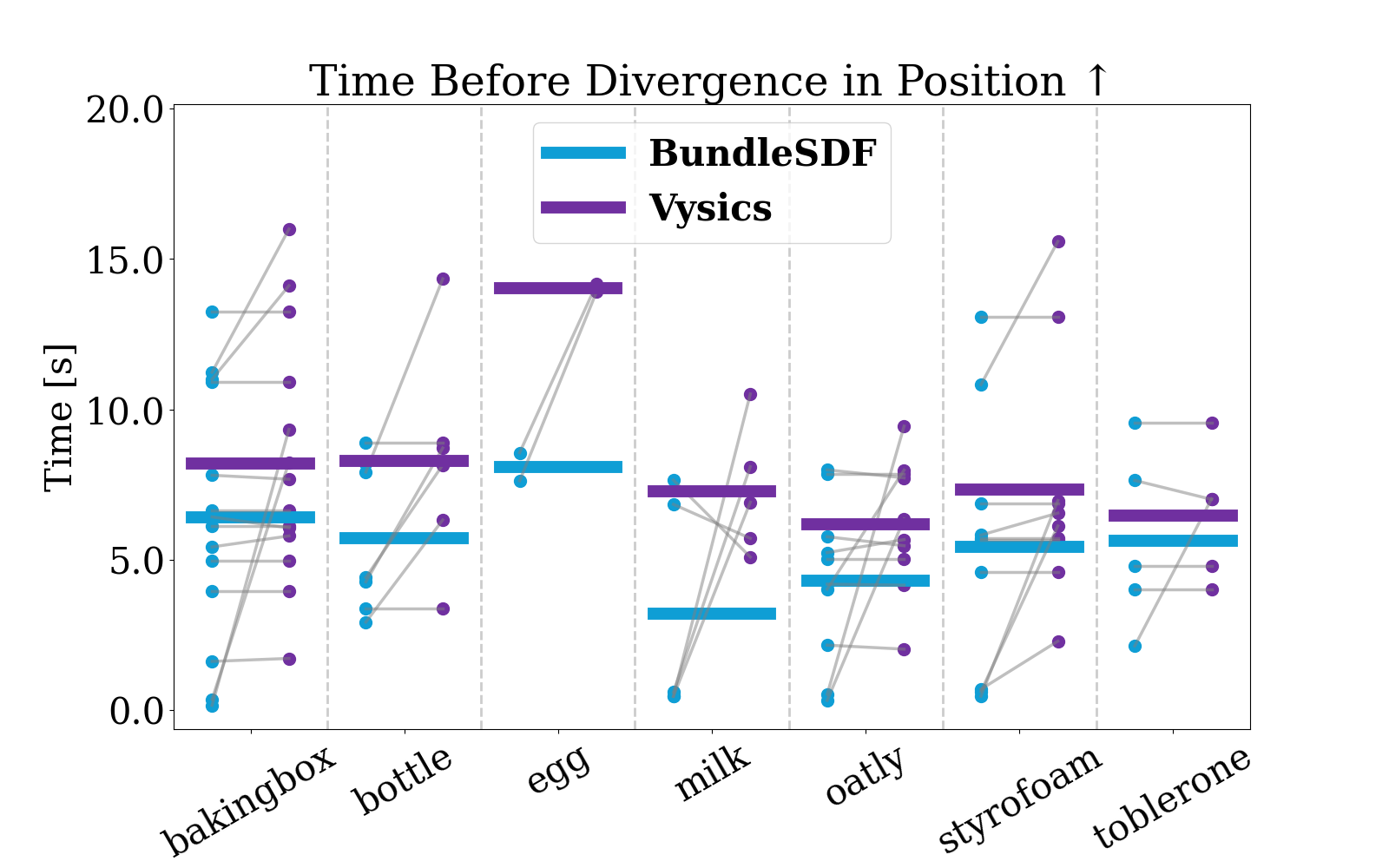}
    \includegraphics[width=\linewidth]{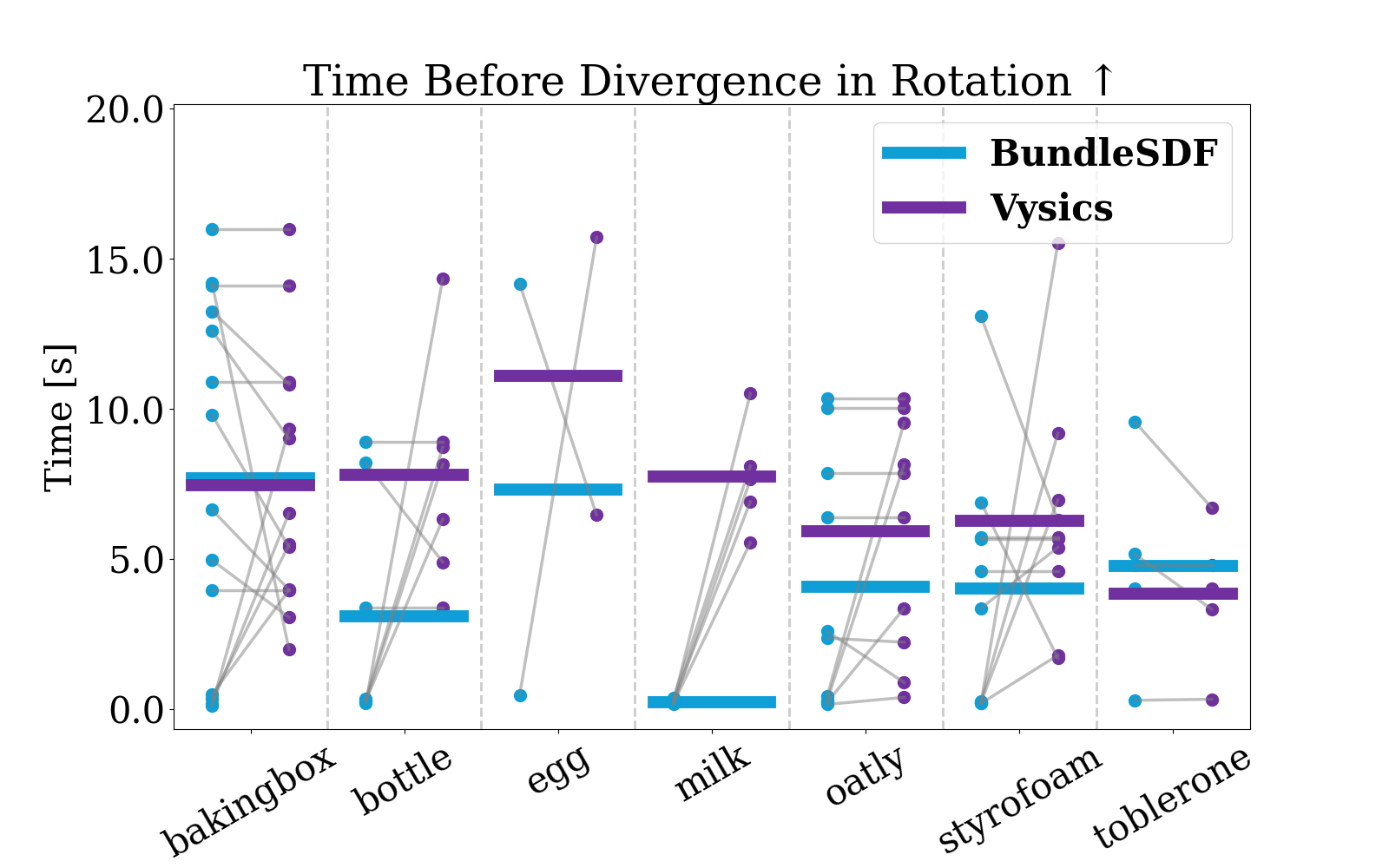}
    \caption{The dynamics prediction accuracy evaluated by the duration of the simulated trajectory under small pose error. }
    \label{fig:dynamics_time_to_drift}
\end{figure}

\begin{figure}[t]
    \centering
    \includegraphics[width=\linewidth]{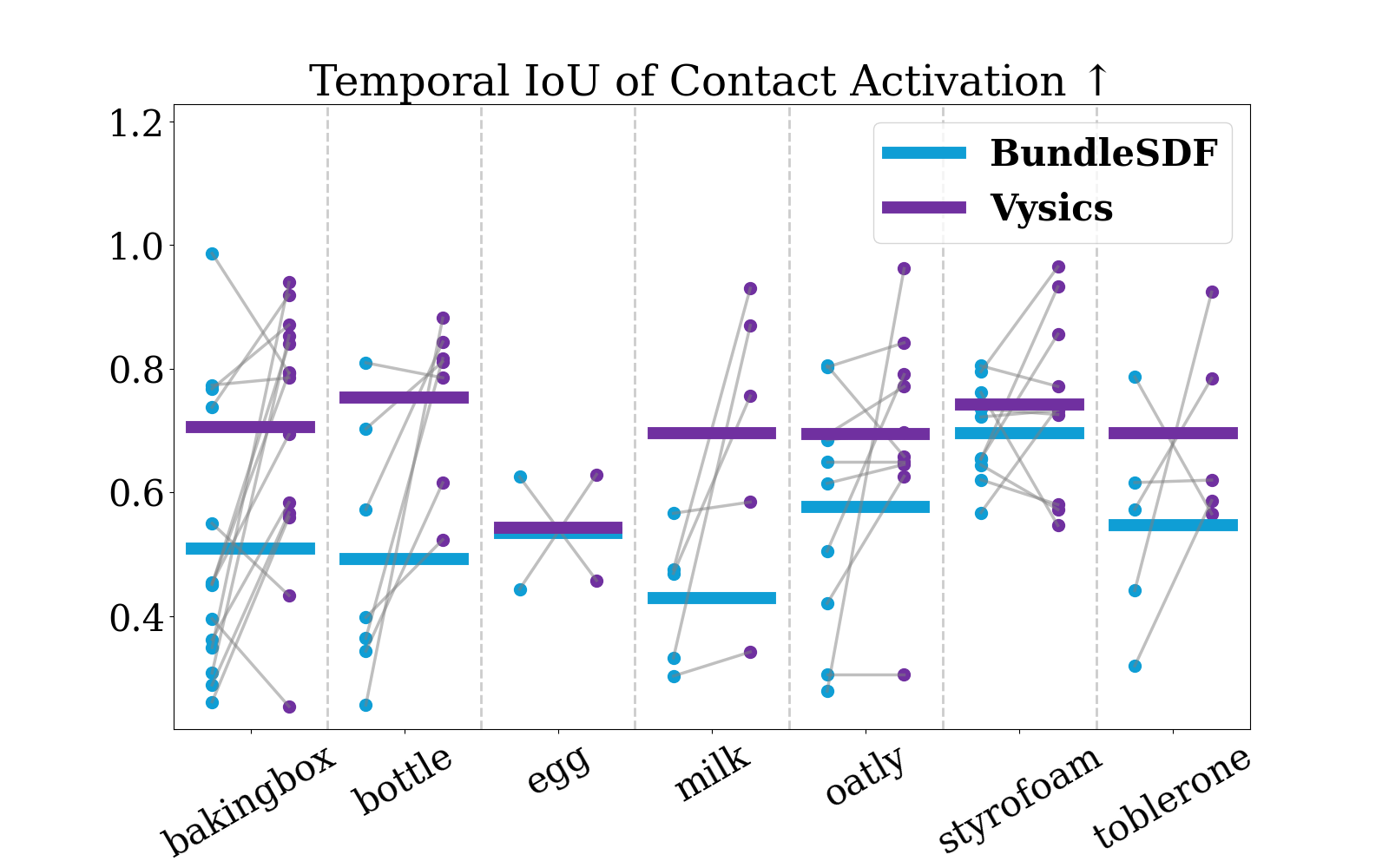}
    \caption{The dynamics prediction accuracy evaluated by the temporal overlap of robot-object contact happening in simulation and in the real world. The real world ground truth was manually annotated.}
    \label{fig:dynamics_contact_iou}
\end{figure}
\section{Results} \label{sec:results}

\subsection{Geometry Reconstruction}
We first compare the geometry reconstruction of our method with that of shape completion models and single-view 3D generation models.  Table~\ref{tab:chamfer_baselines} presents the quantitative results of shape completion models in chamfer distance, averaged per object and over all objects, compared with BundleSDF and Vysics. Under severe occlusion, while the shape completion models can achieve similar or slightly lower chamfer distance than pure vision-based reconstruction, BundleSDF, they fall behind Vysics by a large margin, showing that the data-driven completion models are not as successful as Vysics at filling in the missing pieces. 
Qualitative results of the single-view 3D generation models are shown in \cref{fig:gen_qual}. We find that these generative models typically assume an unobstructed view of the object and do not generate a complete shape when given a partially occluded view. Therefore, we do not evaluate these models quantitatively.

In the following, we provide detailed comparisons between Vysics and BundleSDF, as they both do not require a pretrained model on a large object dataset.
Qualitative comparisons of Vysics and BundleSDF geometric reconstruction is shown in \cref{fig:geometry_qualitative}. In these examples, the robot arm touches an occluded portion of the geometry and causes motion of the object. The view of the object is heavily occluded by the obstacles (books/blanket), resulting in BundleSDF missing a significant portion of the object in its geometry estimate. In contrast, our method recovers the occluded geometry so that the robot arm's interactions with the objects can explain the observed object trajectory.
\cref{fig:geo_quantitative} shows the quantitative results of geometric reconstruction for each individual session in terms of the surface-based metric, chamfer distance, and the volume-based metric, IoU. Our method recovers the occluded geometry through physics-based reasoning over the observed trajectories, substantially and consistently improving the geometric accuracy in both metrics.

\subsection{Dynamics Predictions}

\begin{figure}[t]
    \centering
    \includegraphics[width=\linewidth,trim={0mm 10mm 0mm 0mm},clip]{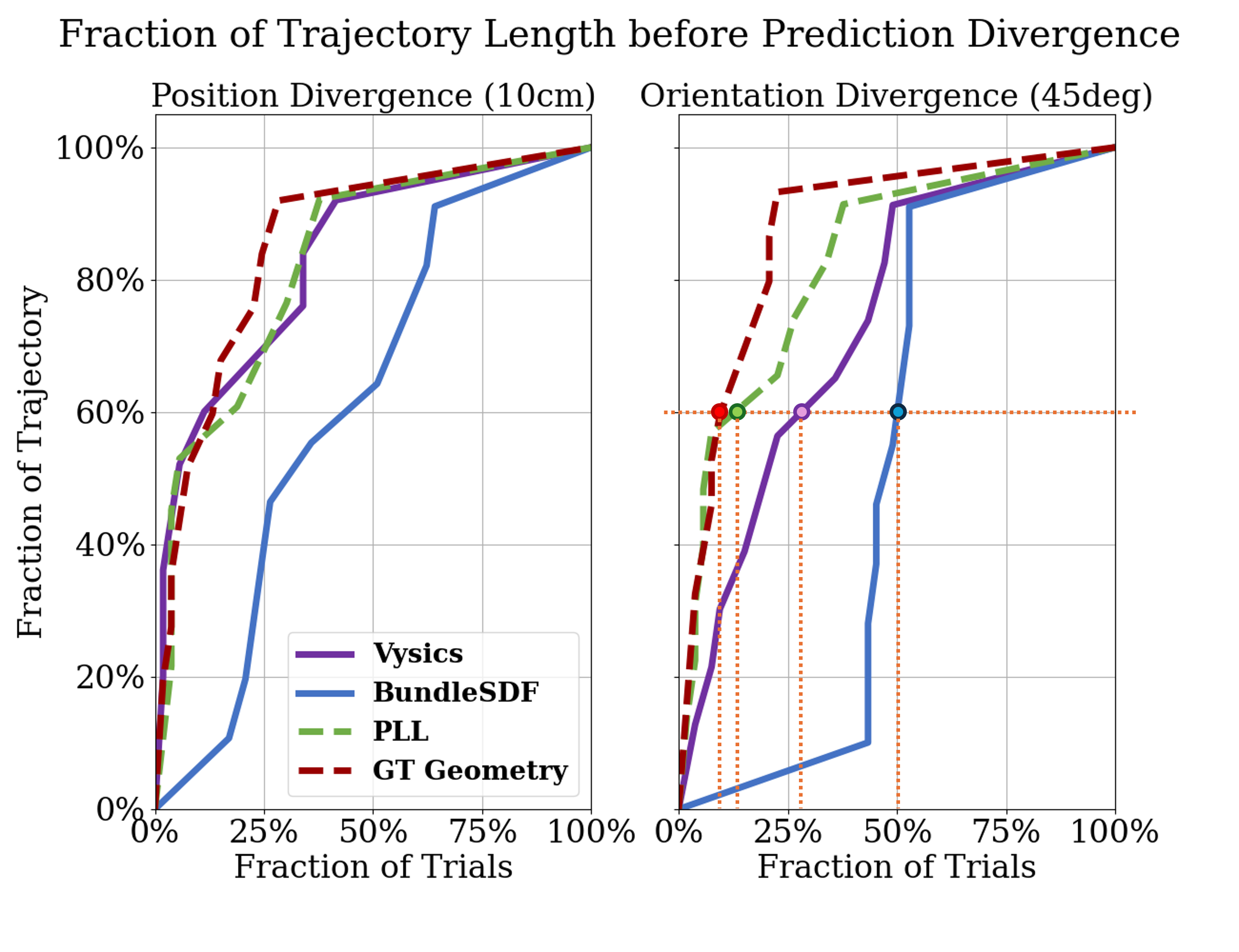}
    \caption{For quantifying dynamics prediction performance, we compare how far into an open-loop rollout the predicted pose stays within 10cm of position error and within 45 degrees of rotational error from the BundleSDF tracked poses.  We normalize the y-axis to the length of the trajectory.  An example interpretation from the right plot (the orange dashed lines):  60\% into the predicted trajectory, approximately half of the BundleSDF dynamics predictions diverged in orientation compared to 30\% of the Vysics dynamics predictions and 10\% of the ground truth geometry and PLL predictions.}
    \label{fig:dynamics_cdf}
\end{figure}

We further use dynamics predictions to show that the geometry estimated by our method better explains the observed trajectory. \cref{fig:dynamics_pose_error} shows the average position and rotation errors when predicting the entire length of the original trajectory as an open-loop rollout.  The trajectories range in length from 3 to 18 seconds. Compared with the shape estimated by the vision-only method, the geometry estimated by our method enabled much smaller simulation pose errors across the dataset.

A similar trend can be found in the time-before-divergence metric in Figures \ref{fig:dynamics_time_to_drift} and \ref{fig:dynamics_cdf}. Our method maintained a small pose error for a longer time than the vision-only baseline in simulations.  Figure \ref{fig:dynamics_cdf} compares Vysics against the vision-only baseline, physics-only baseline, and a baseline featuring a simulation with the ground truth geometry.  As expected, the ground truth geometry simulations maintained more accurate dynamics predictions for the longest.  We point out that even this baseline is imperfect, despite using essentially perfect geometry, due to inaccurate modeling assumptions such as object rigidity and the divergent nature of the dynamics in many of our robot interactions.  Vysics and PLL perform closely to this baseline, though Vysics is moderately worse in orientation divergence.  While most of the dynamics performance by PLL is retained in Vysics, it is unsurprising to see a slight performance drop, given PLL optimizes only for physics accuracy while Vysics balances with visual objectives.  The vision-only baseline is the least performant in both position and orientation rollout accuracy.

The temporal IoU of contact activation is shown in \cref{fig:dynamics_contact_iou}. Our method achieved better temporal consistency between the observed contact and the contact reproduced in the simulations than the vision-based geometry. Despite the chaotic nature of open-loop simulation, our method achieved consistent improvements in all three dynamics prediction metrics across the dataset, which shows the capability of our method in building better object models by exploiting the vision and physics information.


\section{Limitations and Future Work}
\label{sec:limitations}

A limitation of Vysics is that it does not incorporate notions of object elasticity or bounciness into the learning problem.  This shortcoming means the dynamics predictions of our learned models could deviate drastically from the ground-truth trajectories, which saw more chaotic bouncing and slower energy dissipation.  While automated searches for simulator contact parameters exist \cite{acosta2022validating}, preliminary attempts to improve our dynamics prediction accuracy with these methods did not yield significant improvements over our inelastic contact model.

A considerable limitation of our method in its current form is that it often cannot recover from poor pose estimates.  
In our experience, this issue is accentuated by the challenges of our occlusion-rife, contact-rich dataset.  We found shorter video lengths usually resulted in more consistent pose tracking, but this is at odds with the benefits of more data for dynamics parameter regression.  It is possible that the geometry supervision from PLL during BundleSDF's second run can help it perform better pose estimation.  In this case, we could cyclically repeat our BundleSDF-PLL process until both shape and trajectories converge.  
Future versions of Vysics may consider posing a single joint learning problem that performs pose estimation and dynamics model building simultaneously.

The scarcity of observation fundamentally limits the quality of the reconstructed shape from our method. The visual data is heavily occluded, while the contact signal is sparse by nature, leaving limited information for our method to optimize over. Therefore, readers may find the reconstructed shapes qualitatively unappealing, with noisy artifacts and limited details. A potential future direction is to leverage the data-based priors in foundation models to guide shape estimation while respecting visual and physics signals.

Lastly, our experiments featured robot interactions from teleoperation with diversity of interactions in mind.  In a closed-loop system, it could be possible for explorative strategies to determine where the robot should initiate contact to lower its uncertainty about the object's properties.  While this interactive approach is compatible with Vysics as designed, this remains future work.

\section{Conclusion}
\label{sec:conclusion}
Vysics equips robots with the ability to construct high-fidelity dynamics models of novel objects, identified from vision and proprioception, in the face of contact-rich interactions and extremely small amounts of data.  This is the first step toward unifying vision-based geometry estimation with contact dynamics.  
Beyond improvements to the core method of Vysics, discussed in \S \ref{sec:limitations}, future work might replace teleoperated data collection with autonomous, active exploration or the integration of these learned models with planning and control to accomplish some desired task.


\section*{Acknowledgments}
We thank our anonymous reviewers, who provided thorough and fair feedback that improved the quality of our paper.  This work was supported by a National Defense Science and Engineering Graduate (NDSEG) Fellowship, an NSF CAREER Award under Grant No. FRR-2238480, and the RAI Institute.


\bibliographystyle{plainnat}
\bibliography{references}

\clearpage
\appendix
\section{Hyperparameters} \label{apx:hyperparams}
\subsection{BundleSDF Hyperparameters}
\begin{table}[h]
    \centering
    {
    \begin{tabular}{|p{4.5cm}|c|c|}
        \hline
        Description & Symbol in \cite{wen2023bundlesdf} & Value \\ \hline \hline
        RGB loss weight & $w_c$ & 100 \\ \hline
        Uncertain free space loss weight & $w_u$ & 50 \\ \hline
        Uncertain free space SDF target & $\epsilon$ & 0.1 \\ \hline
        Empty space loss weight & $w_e$ & 1 \\ \hline
        Near-surface space loss weight & $w_{surf}$ & 3,000 \\ \hline
        Truncation distance of the near-surface space & N/A & 0.01 \text{m} \\ \hline
        SDF target corresponding to the truncation distance & $\lambda$ & 1 \\ \hline
        Eikonal loss weight & $w_{eik}$ & 0 \\ \hline
        (Ours) Support point loss weight on \eqref{eqn:loss_support_point} & N/A & 2 \\ \hline
        (Ours) Hyperplane-constrained loss weight on \eqref{eqn:loss_hyperplane} & N/A & 1 \\ \hline
        (Ours) Convexity loss weight on \eqref{eqn:loss_convex} & N/A & 1 \\ \hline
    \end{tabular}}
    \caption{Hyperparameters for modified BundleSDF \cite{wen2023bundlesdf}. }
    \label{tab:bsdf_hyperparams}
\end{table}

BundleSDF \cite{wen2023bundlesdf} employed five loss terms to optimize the object SDF and poses: 
\begin{itemize}
    \item RGB loss: a penalty on the discrepancy between the color rendered from the SDF and object pose and the observed RGB value. 
    \item Uncertain free space loss: encourages the SDF to predict a small constant in regions near observed object surfaces.  This preserves the shape's contour while being able to quickly adapt to more reliable observations.
    \item Empty space loss: encourages the foreground empty space in front of the depth readings up to a truncation distance to predict the constant truncation SDF. 
    \item Near-surface loss: for near-surface points with distance to the depth readings under the truncation distance, encourages the SDF to be predicted exactly. 
    \item Eikonal loss: encourages the regressed SDF to have a gradient of magnitude 1 at any point. While described in \cite{wen2023bundlesdf}, this loss term has a \href{https://github.com/NVlabs/BundleSDF/blob/master/config.yml#L84}{weight of 0 in BundleSDF's official implementation}.
\end{itemize}
The weights for the loss terms introduced in Vysics are listed in Table \ref{tab:bsdf_hyperparams}. 

\subsection{PLL Hyperparameters}
\begin{table}[h]
    \centering
    { 
    \begin{tabular}{|l|c|c|}
        \hline
        Description & Symbol in \cite{bianchini2023simultaneous} & Value \\ \hline \hline
        Prediction loss weight & $w_\text{pred}$ & 4 \\ \hline
        Complementarity loss weight & $w_\text{comp}$ & 1 \\ \hline
        Dissipation loss weight & $w_\text{diss}$ & 5000 \\ \hline
        Penetration loss weight & $w_\text{pen}$ & 0.5 \\ \hline
        (Ours) BundleSDF loss weight on \eqref{eqn:vision_loss_actual} & N/A & 0.04 \\ \hline
    \end{tabular}}
    \caption{Loss term weights for modified PLL \cite{bianchini2023simultaneous, pfrommer2020contactnets}.}
    \label{tab:pll_hyperparams}
\end{table}
PLL \cite{bianchini2023simultaneous, pfrommer2020contactnets} features a violation-based implicit loss function with 4 main components:
\begin{itemize}
    \item Prediction loss:  a penalty on the difference between the dynamics model's predicted next state (using hypothesized contact forces) and the observed next state.
    \item Complementarity loss:  a penalty on violation of contact complementarity, i.e. penalize any predicted contact forces if the geometry model does not predict contact.
    \item Dissipation loss:  a penalty on violation of the Coulomb friction model.
    \item Penetration loss:  a penalty on the penetration of the current geometric model into the known environment surface at the end of the time step.
\end{itemize}
PLL learns dynamics models as a bilevel optimization problem:  1) Per epoch, the hypothesized set of contact forces are selected to minimize the sum of the above terms.  2) Between epochs, the dynamics model parameters update via gradient descent.  See \cite{pfrommer2020contactnets} for more details.  We add \eqref{eqn:vision_loss_actual} as a term directly to the above from PLL with the loss weights according to Table \ref{tab:pll_hyperparams}.

\subsection{Other Vysics Hyperparameters}

\subsubsection{Hyperparameters for converting from PLL outputs to modified BundleSDF inputs}
\S \ref{subsec:contact_in_vision_geom} describes querying the PLL geometry for supervising BundleSDF via \eqref{eqn:loss_support_point} and \eqref{eqn:loss_hyperplane}.  Since odometry-based contact learning may lack signal in many directions, data for supervising the next round of BundleSDF is extracted by running all of the input trajectory data again (after PLL has already been trained) through PLL's violation-based implicit loss.  Vysics retains only the queried points with the top fraction of hypothesized contact normal forces -- we used the top 30\%.  Thus, we yield hypothesized-force-filtered data points $\{ (\mathbf{\hat{n}}^p, \mathbf{s}^p)_i \}$ that confidently saw contact during PLL training.

We go from one DSF input/output pair $(\mathbf{\hat{n}}^p, \mathbf{s}^p)$ to many SDF inputs/outputs $\{ (\mathbf{p}, l)_j \}$ subject to \eqref{eqn:loss_support_point} by sampling on the ray $\vec{r}$ defined by $(\mathbf{\hat{n}}^p, \mathbf{s}^p)$:
\begin{itemize}
    \item 100 points randomly sampled up to 5mm from $\mathbf{s}^p$ in the $-\mathbf{\hat{n}}^p$ direction (into the convex hull),
    \item 100 points randomly sampled up to 5mm from $\mathbf{s}^p$ in the $+\mathbf{\hat{n}}^p$ direction, and
    \item 50 points randomly sampled up to 10cm from $\mathbf{s}^p$ in the $+\mathbf{\hat{n}}^p$ direction.
\end{itemize}
Per epoch, BundleSDF randomly picks up to 1K of these SDF input/output pairs. 

For obtaining SDF inputs/lower bounds $\{ (\mathbf{q}, l_\text{min})_k \}$ subject to \eqref{eqn:loss_hyperplane}, we first sample 25K points evenly-spaced on the PLL mesh.  We filter out any points that are more than 1mm away from a supporting hyperplane defined by the hypothesized-force-filtered support points and directions $\{ (\mathbf{\hat{n}}^p, \mathbf{s}^p)_i \}$.  This yields a set of points sampled on the PLL learned mesh plus their associated face's outward normal, $\{ (\mathbf{\hat{m}}, \mathbf{v})_n \}$.  Sampling on the mesh instead of using the support points and directions directly yields supervision over a broader portion of the geometry even after filtering based on supporting hyperplane proximity, since the support points themselves are sparse and can be highly local.  We go from one direction/point pair $\{ (\mathbf{\hat{m}}, \mathbf{v}) \}$ to many SDF inputs/lower bounds $\{ (\mathbf{q}, l_\text{min})_k \}$ by sampling:
\begin{itemize}
    \item 200 points randomly sampled within a cylinder of radius 5cm, starting at $\mathbf{v}$ with its axis extending 5mm along the $-\mathbf{\hat{m}}$ direction,
    \item 100 points randomly sampled within a cylinder of radius 5cm, starting at $\mathbf{v}$ with its axis extending 5mm along the $+\mathbf{\hat{m}}$ direction, and
    \item 100 points randomly sampled within a cylinder of radius 10cm, starting at $\mathbf{v}$ with its axis extending 10cm along the $+\mathbf{\hat{m}}$ direction.
\end{itemize}
Per epoch, BundleSDF randomly picks up to 5K of these input/SDF limit pairs.

\subsubsection{Hyperparameters for converting from BundleSDF outputs to modified PLL inputs}
\S \ref{subsec:vision_in_contact_geom} describes querying the BundleSDF geometry to obtain DSF input/output pairs $\{ (\mathbf{\hat{n}}^v, \mathbf{s}^v)_i \}$ for supervising PLL via \eqref{eqn:vision_loss_actual}.  We query 5,768 approximately evenly-spaced $\mathbf{\hat{n}}^v$ directions and obtain their corresponding support points $\mathbf{s}^v$ on the BundleSDF geometry.  Per epoch, PLL randomly picks up to 1K of these points to incorporate into its loss via \eqref{eqn:vision_loss_actual}.

\end{document}